%% file: neurips_2026.tex
\documentclass{article}

 \usepackage[preprint]{neurips_2026}

\usepackage{float}
\usepackage{xcolor}
\usepackage{amsmath}
\usepackage{amssymb}
\usepackage{mathtools}
\usepackage{amsthm}
\usepackage{algorithm}
\usepackage{threeparttable}
\usepackage{algpseudocode}

\input{math_commands.tex}

\usepackage[utf8]{inputenc} 
\usepackage[T1]{fontenc}    
\usepackage{url}            
\usepackage{booktabs}       
\usepackage{amsfonts}       
\usepackage{nicefrac}       
\usepackage{microtype}      
\usepackage{xcolor}         
\usepackage{wrapfig}
\usepackage{mine}
\usepackage{cleveref}
\usepackage{caption}
\usepackage{minitoc}

\newcommand{\am}[1]{{\color{orange} }}
\newcommand{\aah}[1]{{\color{magenta} }}
\newcommand{\mo}[1]{\textcolor{violet}{}}
\newcommand{\scott}[1]{\textcolor{red}{}}

\definecolor{mcgillred}{RGB}{237, 27, 47}
\newcommand{\dave}[1]{\textcolor{mcgillred}{}}

\newcommand{\hc}[1]{ {}}

\title{Drift Q-Learning}

\author{%
  Anas Houssaini\thanks{Equal contribution. \protect\\ \hspace*{1.8em}Corresponding authors: \texttt{\{achraf.elhoussaini, mo.danesh\}@mail.mcgill.ca}} \quad
  Mohamad H. Danesh\footnotemark[1] \quad
  Amin Abyaneh \\\\
  \textbf{Scott Fujimoto} \quad
  \textbf{Hsiu-Chin Lin} \quad
  \textbf{David Meger} \\\\
  McGill University \\
  Mila - Quebec AI Institute
}

\begin{document}

\maketitle

\begin{abstract}
Offline reinforcement learning requires improving a policy from fixed data while avoiding out-of-distribution actions with unreliable value estimates. Diffusion and flow policies handle this trade-off by modeling the behavior distribution to regularize the RL objective, but they require iterative denoising, solver integrations, and in more efficient variants, distillation or other approximations at inference. We propose \textbf{\ourMethod}, which combines a drift-based behavioral regularizer with critic-driven policy improvement. The value signal biases the policy toward high-value regions of the data support, while attraction and repulsion together keep generated actions near the data and prevent collapse onto a single mode. \ourMethod is implemented as a single network with a unified training objective and generates actions in a \textit{single} forward pass. On D4RL and OGBench, \ourMethod consistently outperforms diffusion and flow methods, advancing the state of the art. Under degraded data quality, where the baselines visibly struggle, \ourMethod remains close to its clean-data performance, positioning it as a promising alternative to diffusion and flow-based methods while maintaining the simplicity and efficiency of deterministic approaches.
\begin{center}
\textbf{Project page:} \href{https://driftql.github.io/}{driftql.github.io}
\end{center}
\end{abstract}

\section{Introduction}\label{sec:intro}

Offline reinforcement learning (RL) learns policies from fixed datasets without environment interaction \citep{fujimoto2019bcq}. Since the value function is trained exclusively on transitions from the dataset, its estimates for out-of-distribution (OOD) actions are unreliable. This necessitates behavior regularization that must balance two competing objectives: constraining the policy toward the data distribution so that value estimates remain trustworthy while still permitting policy improvement away from low-value actions.

Expressive generative models~\citep{sohl2015deep_diffusion_original, Lipman2022FlowMF} are well-suited for this tradeoff as they can represent the full multimodal structure of the behavior distribution, providing broad coverage of the data support while remaining flexible enough to concentrate probability on high-value regions. These properties have driven strong empirical performance across offline RL benchmarks \citep{janner2022planning_diffusion, hansen2023idql, diffusionQL, fql_park2025, abyaneh2026contractive}. However, existing approaches carry significant practical limitations. Such models require iterative denoising or heavy solver integrations to produce an action, making them slow at inference time. While distillation-based alternatives can improve inference speed, they require an additional network and use a two-stage training pipeline, increasing complexity.

\dave{These next two paragraphs must be critically tied to the technical contributions in the same language we end up using in Sec 4. Just tagging here while Sec 4 is being re-written.}
Drifting Models offer a promising alternative \citep{driftmodels}. They match the expressive capacity of diffusion and flow methods but generate samples in a single forward pass at both training and inference~\citep{lai2026unified}.
A Drifting Model learns a direct one-step pushforward map from prior noise and conditioning signals to data, supervised by the non-parametric \textit{drifting field} which balances support for data while resisting collapse.

\begin{figure}
    \centering
    \includegraphics[width=1\linewidth]{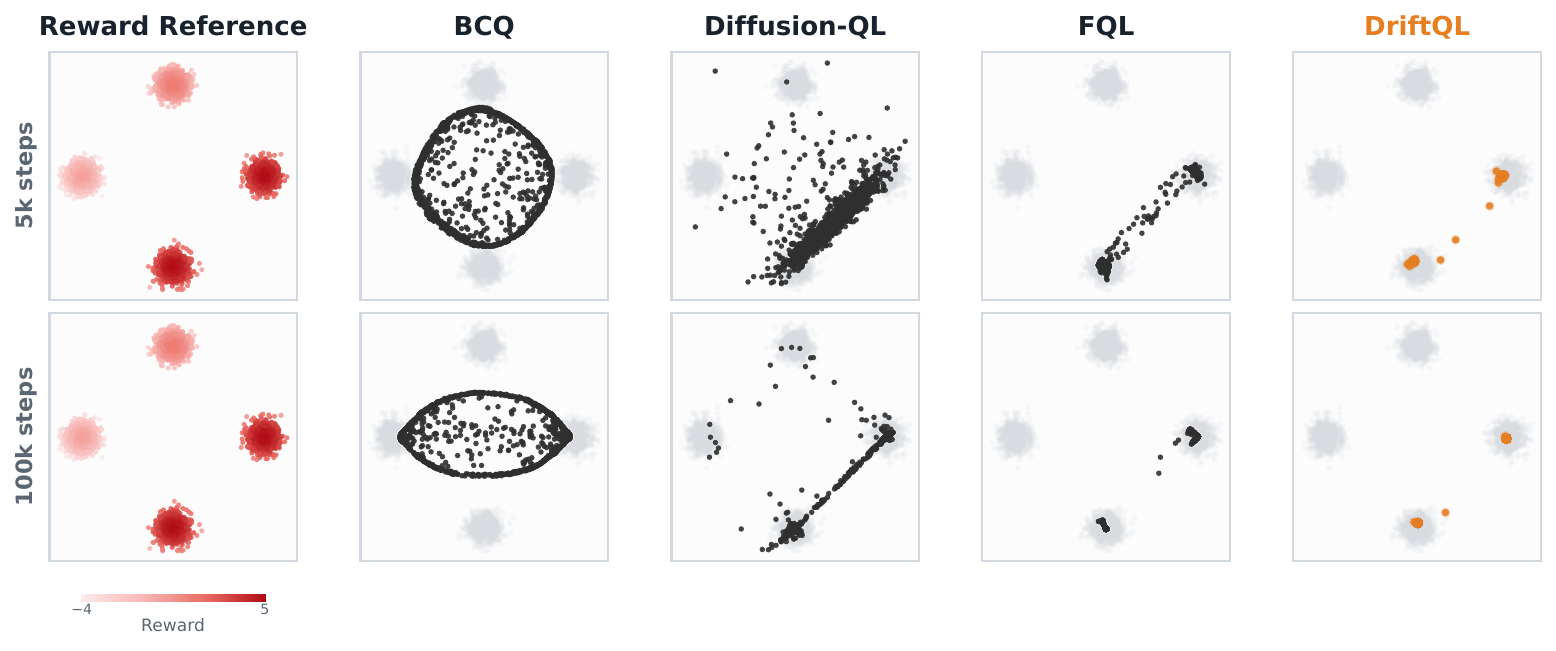}
    \caption{\textbf{Value-guided mode selection on a four-Gaussian bandit with tied optima.} We adapt the four-Gaussian bandit of \citet{diffusionQL}, arranging four isotropic data modes in a cross with the right and bottom modes tied for highest reward. Reward decays with distance from each mode center. The first column shows the reward reference, and gray blobs show dataset support. Rows show policy samples after $5$k and $100$k training steps. BCQ remains broad and assigns mass to unsupported regions, while Diffusion-QL and FQL move toward high-reward modes but retain scattered samples or bridges between modes. \ourMethod concentrates on the two tied high-reward supported modes, with sharper mode selection by $100$k steps.}

\label{fig:intuition_bandit}
\end{figure}

\hc{the figure is not referred. also, what are we trying to show here? I don't see theoretical reasons why driftQL should converge faster}

Building on this insight, we propose Drift Q-Learning (\textbf{\ourMethod}), a one-step generative method
with a conditional drift field \citep{driftmodels} to supervise behavior regularization in offline RL. Drifting models are particularly well-suited to this setting because they reconcile the two requirements identified above: they preserve mass across candidates during training to smooth optimization, while remaining mode-seeking at convergence \citep{lai2026unified} so that probability concentrates on high-value regions rather than the full action distribution. Concretely, our drift field includes a behavior-cloning-like attraction toward observed dataset actions, 
and additionally repels nearby generated actions from one another, preserving a diverse set of candidates within high-value regions throughout training. Finally, since the drift field defines a single-step transport target, inference requires only one forward pass, avoiding both the discretization and integration errors as well as the computational overhead of iterative denoising in diffusion and flow-based generative modeling. \autoref{fig:intuition_bandit} highlights convergence behavior of \ourMethod.

We evaluate \ourMethod on D4RL and OGBench \citep{fu2021d4rl, ogbench_park2025}, where it outperforms diffusion and flow-based baselines while generating actions in a single forward pass with no denoising chains, no solvers, no distillation, and no auxiliary networks. When data quality degrades, \ourMethod sustains its performance in various environments and noise levels where other baselines clearly struggle. Our results position \ourMethod as a promising alternative to diffusion and flow-based offline RL methods while maintaining the simplicity and efficiency of deterministic approaches.

\section{Related Work}
\label{sec:related_work}

\textbf{Offline RL} must improve a policy from fixed data without exploiting unsupported actions whose values are unreliable \citep{fujimoto2019bcq}. Early methods addressed this through explicit behavior constraints \citep{fujimoto2019bcq, kumar2019bear}, while CQL \citep{kumar2020cql} regularizes the critic to suppress overestimated values on OOD actions. IQL \citep{kostrikov2021iql} avoids evaluating OOD actions entirely through expectile regression, and uncertainty-aware methods use critic ensembles to quantify uncertainty and moderate risky policy updates \citep{an2021uncertaintybased, ghasemipour2022so, yang2022rorl, danesh2025safe}. \am{@Moe this section is a bit thin for offline RL recap, also add a sentence on how drifting is different than these offline RL methods to the end (contrastive language), pull some refs from our previous papers or FQL maybe idk.}

\textbf{Generative policies for offline RL} have largely started with diffusion-based methods that combine expressive behavior modeling with value guidance: DQL~\citep{diffusionQL} jointly trains a diffusion policy with behavior cloning and value maximization, IDQL~\citep{hansen2023idql} couples a diffusion behavior model with implicit Q-learning for policy extraction, EDP~\citep{kang2023efficientDP} approximates actions from corrupted samples to avoid running the full denoising chain at training, and BDPO~\citep{gao2025behaviorregularized} derives an analytic KL regularization along the diffusion trajectory within a two-time-scale actor-critic. Flow-based methods further reduce training and inference cost by avoiding iterative sampling: FQL~\citep{fql_park2025} trains an expressive flow-matching policy and distills it into an RL-optimized one-step student, FlowQ~\citep{alles2025flowq} bakes energy-based Q-guidance directly into the flow-matching objective via reweighted regression, and SSCP~\citep{koirala2025flow} augments flow matching to predict direct completion vectors for one-shot action generation. Later extensions further accelerate generation through distillation, flow reformulation, or direct one-step training~\citep{chae2026flow, zhang2026reform, nguyen2026onestep}. \ourMethod takes an entirely different route. We train a one-step policy with drifting loss, sidestepping iterative denoising and reliance on a multi-step teacher altogether.

\section{Background}
\label{sec:background}

\textbf{Offline RL.}
Consider a Markov Decision Process defined by the tuple $\mathcal{M} = \langle \mathcal{S}, \mathcal{A}, P, r, \gamma \rangle$, where $\mathcal{S}$ and $\mathcal{A}$ denote the state and action spaces (with $s, s' \in \mathcal{S}$ and $a \in \mathcal{A}$), $P(s' \mid s, a)$ is the transition function, $r(s, a)$ is the reward, and $\gamma \in [0, 1)$ is the discount factor. An agent behaves according to a policy $\pi(a \mid s)$, aiming to maximize the expected discounted return $J(\pi) = \mathbb{E}_{\tau \sim p_\pi}\left[\sum_{t=0}^{\infty} \gamma^t r(s_t, a_t)\right]$, under trajectories $\tau$ induced by $\pi$ and $P$~\citep{sutton1998reinforcement}. In offline RL, the agent must optimize $J(\pi)$ using only a static dataset $\mathcal{D}=\{(s_i, a_i, r_i, s'_i)\}_{i=1}^M$ collected by an unknown behavior policy $\pi_\beta$ \citep{fujimoto2019bcq}.

Value-based offline RL algorithms estimate the action-value function $Q^\pi(s, a)$, but suffer from \textit{extrapolation error} on OOD actions. Behavioral regularization addresses this by constraining the learned policy to remain close to the behavior policy \citep{fujimoto2021td3bc, kumar2020cql}. A prominent example of this paradigm is TD3+BC \citep{fujimoto2021td3bc}, which augments the standard Q-maximization objective with a BC penalty. The corresponding actor loss is formulated as:
\begin{equation}
    \mathcal{L}_{\text{actor}}(\theta) = -\lambda\, \mathbb{E}_{(s,a)\sim\mathcal{D}}\!\left[Q(s, \pi_\theta(s))\right] + \mathbb{E}_{(s,a)\sim\mathcal{D}}\!\left[\|\pi_\theta(s) - a\|^2\right],
    \label{eq:td3bc}
\end{equation}
where $\lambda = \alpha / \mathbb{E}_{(s,a)\sim\mathcal{D}}[|Q(s,a)|]$ normalizes the Q-value gradients to ensure they are properly scaled relative to the BC penalty.
A downside to TD3+BC is that $\pi_\theta$ is deterministic, meaning that the regularizer produces a single target per state and cannot maintain mass across multiple candidate actions, a property \ourMethod recovers through stochastic generation and a repulsive component.

\textbf{Generative Drifting Models.} Drifting models~\citep{driftmodels} train a one-step conditional generator by constructing a kernel-based vector field that displaces generated outputs toward a target distribution. Let $f_\theta(c,\epsilon)$ map a conditioning input $c$ and Gaussian noise $\epsilon\sim\mathcal{N}(0,I)$ to a generated output $\hat{y}$. For a fixed $c$, independent noise draws induce the generator distribution $q_\theta$, while $p$ denotes the target distribution for that condition.

To compute the drift, positives $y^+\sim p$ are drawn from the target distribution and negatives $\hat{y}^-\sim q_\theta$ are produced by the current generator using random noise. The drift decomposes as:
\begin{align}
V_{p,q_\theta}(\hat{y}) ={}& {\color{myblue}V_p^+(\hat{y})} - {\color{myred}V_{q_\theta}^-(\hat{y})},\\
& {\color{myblue}V_p^+(\hat{y})} = \frac{1}{Z_p(\hat{y})}\mathbb{E}_{y^+\sim p}\left[k(\hat{y},y^+)(y^+-\hat{y})\right],\label{eq:pos_drift}\\
& {\color{myred}V_{q_\theta}^-(\hat{y})} = \frac{1}{Z_q(\hat{y})}\mathbb{E}_{\hat{y}^-\sim q_\theta}\left[k(\hat{y},\hat{y}^-)(\hat{y}^- - \hat{y})\right].\label{eq:neg_drift}
\end{align}
Here $k(\cdot,\cdot)$ is a similarity kernel with $Z_p(\cdot)$ and $Z_q(\cdot)$ normalizing the kernel weights. Both terms take the form of a mean-shift step, which moves a point toward the kernel-weighted average of a reference set, giving closer samples stronger influence. $V^+_p$ applies this to the positives, attracting $\hat{y}$ toward target outputs. $V^-_{q_\theta}$ does the same over generated samples, and the minus sign converts the resulting pull into a push away~\citep{lai2026unified}. When $q_\theta = p$, the two terms cancel. In practice, the expectations are estimated with $M$ positives $\{y_i^+\}_{i=1}^M$ from $p$ and $N$ negatives $\{\hat{y}_j^-\}_{j=1}^N$ from the generator, where $\hat{y}_j^- = f_\theta(c, \epsilon_j^-)$ for independent $\epsilon_j^- \sim \mathcal{N}(0,I)$. The generator is trained by regressing each output toward its drifted location:
\begin{equation}
\mathcal{L}_{\rm drift}(\theta)
=\mathbb{E}_{c,\epsilon}\left[\left\|\hat{y} - \operatorname{sg}\left(\hat{y} + V_{p,q_\theta}(\hat{y})\right)\right\|_2^2\right], \qquad \hat{y} = f_\theta(c,\epsilon),
\end{equation}
where $\operatorname{sg}(\cdot)$ denotes stop-gradient. After training, inference is a single evaluation $\hat{y} = f_\theta(c, \epsilon)$.
\hc{ $\hat{y} = f_\theta(c, \epsilon)$ was mentioned 3 times here. necessary?}

\section{Drift Q-Learning}\label{sec:driftql}

\begin{figure}[t]
    \centering
    \includegraphics[width=\linewidth]{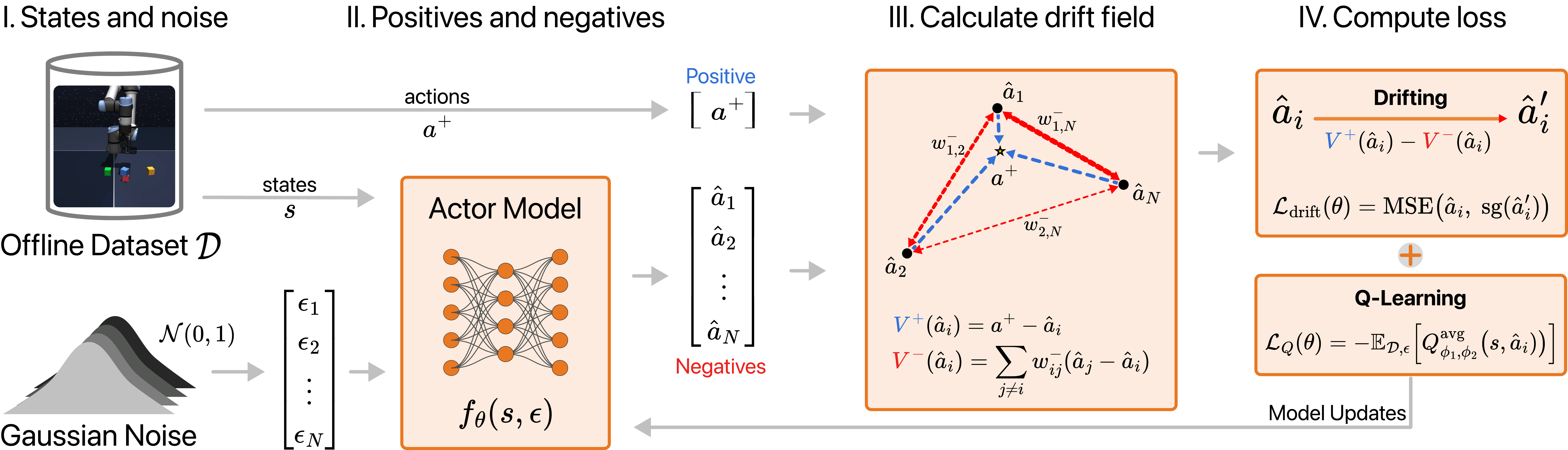}
    \caption{\textbf{Overview of \ourMethod.} (I) A state $s$ is sampled from the offline dataset $\mathcal{D}$ along with $N$ Gaussian noise vectors $\epsilon$. (II) The actor network processes state and noise to generate candidate actions $\{\hat{a}_i\}_1^N$. (III) The generated actions are subjected to a conditional drift field composed of two forces: \textcolor{myblue}{an attraction force ($V^+$)} that pulls all actions uniformly toward the true dataset action $a^+$, and \textcolor{myred}{a repulsion force ($V^-$)} that pushes the generated actions away from one another. (IV) At the end, the drifting process dictates the update step, computing a new target action $\hat{a}'_i$ for each sample, calculating the drift loss, and updating the actor network with both value estimation and drifting losses.}
    \label{fig:overall}
\end{figure}

In this section, we present \ourMethod, a behavior-regularized offline RL learning method that builds on the drifting framework.
We adopt the one-step pushforward distribution in state-conditioned action space as our policy parameterization. Due to the nature of data in offline RL, constructing the supervising conditional drift field in this situation requires a critical set of changes compared to the original formulation~\citep{driftmodels}. An actor-critic objective unifies distribution modeling with reward-seeking and enables the learning of effective behaviors from challenging offline data. \ourMethod's logic is outlined in \autoref{fig:overall}.

\hc{III. the weights should be $w_{1,:}^-, w_{2,:}^-, w_{N,:}^-$. IV. the equations are not consistent with the main text. Just put $L_{drift}$}
\hc{figure is referred?}

At its core, \ourMethod uses a stochastic generator $a = f_\theta(s, \epsilon)$ as the policy. A state $s \in \mathcal{S}$ and a noise vector $\epsilon \sim \mathcal{N}(0, I)$ are mapped to an action $a \in \mathcal{A}$, with $\epsilon$ inducing the policy's stochasticity. For each state, the actor produces $N$ candidate actions $\hat{a}_i = f_\theta(s, \epsilon_i)$, $i = 1, \dots, N$, which are simultaneously transported by a state-conditional drift field:
\begin{equation}
  V(\hat{a}_i) = {\color{myblue}V^+(\hat{a}_i)} - {\color{myred}V^-(\hat{a}_i)}.
  \label{eq:drift}
\end{equation}
\textbf{Attraction.} In the unconditional drifting field of \citet{driftmodels}, $V^+_p$ is a kernel-weighted mean shift over multiple samples drawn from the target distribution, with each positive weighted by its similarity to the candidate (\autoref{eq:pos_drift}). However, continuous state and action offline RL is single-positive by construction: the dataset supplies one observed action per state. With a single positive, \autoref{eq:pos_drift} contains only one target sample and the empirical estimator of $V^+_p$ reduces to a single term:
\begin{equation}
  {\color{myblue}V^+(\hat{a}_i)} = a^+ - \hat{a}_i.
  \label{eq:attraction}
\end{equation}
\textbf{Repulsion.} Repulsion retains the mean-shift structure of $V^-_{q_\theta}$ from the original drifting method (\autoref{eq:neg_drift}): the $N-1$ other policy samples form a non-degenerate empirical estimator of the model distribution $q_\theta(\cdot \mid s)$, and the resulting force prevents candidates from collapsing onto $a^+$. Concretely, the repulsion vector is a weighted average of displacements toward neighboring samples, which the minus sign in \autoref{eq:drift} converts into a push away from them:
\begin{equation}
  {\color{myred}V^-(\hat{a}_i)} = \sum_{k \neq i} {\color{myred}w^-_{ik}}\,(\hat{a}_k - \hat{a}_i),
  \label{eq:repulsion}
\end{equation}
where the weights are obtained by a row-wise softmax over the off-diagonal entries of an $N \times N$ logit matrix,
\begin{equation}
  {\color{myred}w^-_{i,:}} = \operatorname{softmax}\bigl({\color{myred}\ell^-_{i,:}}\bigr),
  \label{eq:softmax}
\end{equation}
whose logits come from a Gaussian kernel with temperature $\tau$ controlling the sharpness of the fall-off:
\begin{equation}
  {\color{myred}\ell^-_{ik}} = -\frac{\|\hat{a}_i - \hat{a}_k\|_2^2}{2\tau^2 \, d_a}.
  \label{eq:kernel}
\end{equation}

\textbf{Drift loss.}
Combining the attraction and repulsion components, each generated action $\hat{a}_i$ is transported toward a drifted candidate $\hat{a}^+_i = \hat{a}_i + V(\hat{a}_i)$, where $V(\hat{a}_i)$ is evaluated over the jointly generated set $\{\hat{a}_j\}_{j=1}^N$ for state $s$. The drift loss regresses each $\hat{a}_i$ toward this target, held fixed via a stop-gradient:
\begin{equation}
  \mathcal{L}_{\text{drift}}(\theta) = \mathbb{E}_{s \sim \mathcal{D},\, \epsilon_{1:N}}\!\left[\frac{1}{N}\sum_{i=1}^N \bigl\|\hat{a}_i - \operatorname{sg}\!\bigl(\!\operatorname{clip}(\hat{a}^+_i,\, -1, 1)\bigr)\bigr\|_2^2\right], \quad \hat{a}^+_i = \hat{a}_i + V(\hat{a}_i).
  \label{eq:drift_loss}
\end{equation}
The stop-gradient ($\operatorname{sg}$) freezes the drifted target $\hat{a}^+_i$, so the actor is updated toward the transported point without backpropagating through the drift-field computation. Since the targets are recomputed from the current policy at every training step, the drift regularizer tracks the evolving actor distribution.

\textbf{Critic.} Following FQL \citep{fql_park2025}, we pair the drift actor with a clipped double-Q critic $\{Q_{\phi_1}, Q_{\phi_2}\}$ \citep{fujimoto2018addressing}, trained via standard Bellman regression.

\textbf{Actor.} The actor jointly minimizes the drift loss and a value-improvement term. The drift loss acts as a behavior regularizer, while the value gradient pushes mass toward high-value modes.
\begin{equation}
  \mathcal{L}_{\text{actor}}(\theta) = \alpha\,\mathcal{L}_{\text{drift}}(\theta) + \mathcal{L}_Q(\theta),
  \qquad
  \mathcal{L}_Q(\theta) = -\mathbb{E}_{\mathcal{D},\epsilon}\Bigl[\tfrac{1}{2}\bigl(Q_{\phi_1} + Q_{\phi_2}\bigr)\bigl(s, f_\theta(s, \epsilon)\bigr)\Bigr],
  \label{eq:actor_loss}
\end{equation}
with $\alpha > 0$ weighting behavioral fidelity against policy improvement. The value gradient $\nabla_\theta \mathcal{L}_Q$ supplies per-candidate reweighting.
Candidates are not merely pulled toward $a^+$, but are biased toward high-value regions of the support, allowing the policy to concentrate on multiple high-value modes when the data contains them.

The drift policy is trained directly against \autoref{eq:actor_loss} and, unlike competing generative model approaches~\citep{diffusionQL, fql_park2025}, inference reduces to a single forward pass $a = f_\theta(s, \epsilon)$. With no need for an auxiliary network, solvers, or denoising chain, we argue that this formulation is the most direct implementation of generative behavior regularization for offline RL policy improvement. Its efficiency is empirically validated in \autoref{sec:exp_latency}.

\textbf{Differences from \citet{driftmodels}.} The original drifting implementation was developed for high-dimensional, unconditional image generation, where the drift field must single-handedly shape the model distribution toward the data. Two structural challenges arise in that setting, both highlighted by \citet{lai2026unified}. First, when kernels are evaluated on learned feature embeddings, relative distances among samples can become nearly uniform, making the softmax kernel weights close to flat and the drift magnitude small even when $p$ and $q$ remain mismatched. Second, the Laplace kernel (when replaced with the Gaussian one) does not give a clean identifiability story: the preconditioned-score decomposition shows that mean-shift differs from the smoothed-score field by a scalar preconditioner and a covariance residual, both depending on the local kernel-reweighted neighborhood, so the equilibrium $V \equiv 0$ no longer forces $q = p$ unless those terms are separately controlled.

To compensate, \citet{driftmodels} compute a \textit{joint} kernel coupling positives and negatives, $V_{p,q}(x) \propto \mathbb{E}[k(x, y^+)\,k(x, y^-)\,(y^+ - y^-)]$, apply softmax along both the $x$ and $y$ axes, aggregate drift fields across multiple temperatures, and rescale the resulting force to unit RMS. These mechanisms reshape the effective transport field so that, even under feature-space evaluation with a Laplace kernel, the drift remains a sufficient training signal on its own.

In unconditional image generation, the drift field is the \textit{only} signal shaping the generator, so it must single-handedly drive the model to a unique, identifiable equilibrium $q=p$. This is precisely the job of the symmetrized affinities, cross-weighting, multi-temperature aggregation, and force normalization. \citet{lai2026unified} show that without them the bare kernel drift does not guarantee $V \equiv 0 \Rightarrow q=p$, so the machinery exists to suppress that residual ambiguity.

Offline RL removes the need for it on two grounds, decided \textit{before} any tuning. First, we do not target $q=p$ at all. The optimal policy is generally deterministic, and the goal is to \textit{improve} over the behavior policy and concentrate on high-value modes, not to reproduce the data distribution.

Second, and more importantly, the drift is no longer the only signal. The value gradient $\nabla_\theta \mathcal{L}_Q$ supplies an independent, dataset-anchored pull toward high-value support that anchors the actor regardless of the drift's internal balance. The roles therefore separate cleanly, with the drift regularizer keeping generated actions near dataset support while the critic supplies the value-guided improvement signal, so the coefficient $\alpha$ in \autoref{eq:actor_loss} directly governs the trade-off between the two. We thus expect the drift to remain stable in \ourMethod without the image-oriented machinery, and verify this in \aref{app:contragen}. The simplified drift computation matches or improves on the original drifting implementation in the offline RL setting, and a stress test that deliberately breaks the drift's attraction-repulsion balance confirms that the critic, not the machinery, supplies the missing constraint.

\section{Experiments}
\label{sec:experiments}

We evaluate \ourMethod on a suite of standard offline RL benchmarks designed to test behavioral multi-modality, long-horizon planning, and distributional coverage. Our experiments are structured to comprehensively validate both the downstream performance and the underlying mechanics of our approach. Specifically, we investigate whether \ourMethod matches the expressivity of iterative generative policies, discuss computational efficiency during training and inference, and isolate the impact of our drift field adaptations.

\subsection{Experimental Setup}

\textbf{Benchmarks.} We evaluate our approach on two primary benchmark suites. To test standard continuous control and navigation, we use AntMaze, Adroit, and Locomotion from D4RL \citep{fu2021d4rl}. To specifically stress-test the policy's ability to handle highly multimodal and suboptimal data distributions with sparse reward signals, we evaluate on the OGBench suite \citep{ogbench_park2025}.

\textbf{Baselines.} We compare \ourMethod against a comprehensive set of offline RL algorithms. To represent standard Gaussian policies, we evaluate against Behavioral Cloning (BC), Implicit Q-Learning (IQL) \citep{kostrikov2021iql}, and ReBRAC \citep{tarasov2023revisiting}. For expressive generative policies, we benchmark against a diverse suite of diffusion and flow-matching methods. Our diffusion baselines include IDQL \citep{hansen2023idql}, SRPO \citep{chen2024score}, and CAC \citep{ding2024consistency}, and the flow-based ones are FQL \citep{fql_park2025} and its associated flow-based variants: Implicit Flow Q-Learning (IFQL), Flow Advantage-Weighted Actor-Critic (FAWAC), and Flow Behavior-Regularized Actor-Critic (FBRAC) \citep{fql_park2025}.

\begin{table*}
\caption{
\textbf{Offline RL results.} \ourMethod achieves competitive or superior performance in almost all 78 offline learning tasks. We present standard deviations after ``$\pm$'' and denote values at or above 95\% of the best performance in bold.}
\label{table:offline}
\centering
\resizebox{\textwidth}{!}{
\setlength{\tabcolsep}{6pt}
\begin{threeparttable}
\begin{tabular}{lccccccccccc}
\toprule
\multicolumn{1}{c}{} & \multicolumn{3}{c}{\texttt{Gaussian}} & \multicolumn{3}{c}{\texttt{Diffusion}} & \multicolumn{4}{c}{\texttt{Flow}} & \multicolumn{1}{c}{\texttt{Drift}} \\
\cmidrule(lr){2-4} \cmidrule(lr){5-7} \cmidrule(lr){8-11} \cmidrule(lr){12-12}
\texttt{Task} & \texttt{BC} & \texttt{IQL} & \texttt{ReBRAC} & \texttt{IDQL} & \texttt{SRPO} & \texttt{CAC} & \texttt{FAWAC} & \texttt{FBRAC} & \texttt{IFQL} & \texttt{FQL} & \texttt{\color{myorange}\textbf{\ourMethod}} \\
\midrule
\multicolumn{12}{l}{\textit{D4RL}} \\[2pt]
\texttt{antmaze ($\mathbf{6}$)} & $17$ & $57$ & $78$ & $79$ & $74$ & $30$ {\color{gray}{\tiny $\pm 3$}} & $44$ {\color{gray}{\tiny $\pm 3$}} & $64$ {\color{gray}{\tiny $\pm 7$}} & $65$ {\color{gray}{\tiny $\pm 7$}} & $\mathbf{84}$ {\color{gray}{\tiny $\pm 3$}} &  $\mathbf{84}$ {\color{gray}{\tiny $\pm 9$}} \\
\texttt{adroit ($\mathbf{12}$)} & $48$ & $53$ & $\mathbf{59}$ & $52$ {\color{gray}{\tiny $\pm 1$}} & $51$ {\color{gray}{\tiny $\pm 1$}} & $43$ {\color{gray}{\tiny $\pm 2$}} & $48$ {\color{gray}{\tiny $\pm 1$}} & $50$ {\color{gray}{\tiny $\pm 2$}} & $52$ {\color{gray}{\tiny $\pm 1$}} & $52$ {\color{gray}{\tiny $\pm 1$}} & $50$ {\color{gray}{\tiny $\pm 5$}} \\
\texttt{locomotion ($\mathbf{9}$)} & $50$ & $82$ & $\mathbf{90}$ & $82$ & $\mathbf{87}$ & $-$ & $-$ & $-$ & $50$ {\color{gray}{\tiny $\pm 3$}} & $63$ {\color{gray}{\tiny $\pm 2$}} & $81$ {\color{gray}{\tiny $\pm 23$}} \\
\addlinespace
\cmidrule(lr){2-12}
\textbf{\texttt{D4RL Overall}} & $38$ & $64$ & $\mathbf{76}$ & $71$ {\color{gray}{\tiny $\pm 1$}} & $71$ {\color{gray}{\tiny $\pm 1$}} & $-$ & $-$ & $-$ & $56$ {\color{gray}{\tiny $\pm 9$}} & $66$ {\color{gray}{\tiny $\pm 1$}} & $\mathbf{72}$ {\color{gray}{\tiny $\pm 2$}} \\
\midrule
\multicolumn{12}{l}{\textit{OGBench}} \\[2pt]
\texttt{antmaze-large-st ($\mathbf{5}$)} & $11$ {\color{gray}{\tiny $\pm 1$}} & $53$ {\color{gray}{\tiny $\pm 3$}} & $81$ {\color{gray}{\tiny $\pm 5$}} & $21$ {\color{gray}{\tiny $\pm 5$}} & $11$ {\color{gray}{\tiny $\pm 4$}} & $33$ {\color{gray}{\tiny $\pm 4$}} & $6$ {\color{gray}{\tiny $\pm 1$}} & $60$ {\color{gray}{\tiny $\pm 6$}} & $28$ {\color{gray}{\tiny $\pm 5$}} & $79$ {\color{gray}{\tiny $\pm 3$}} & $\mathbf{92}$ {\color{gray}{\tiny $\pm 4$}}\\
\texttt{antmaze-giant-st ($\mathbf{5}$)} & $0$ {\color{gray}{\tiny $\pm 0$}} & $4$ {\color{gray}{\tiny $\pm 1$}} & $26$ {\color{gray}{\tiny $\pm 8$}} & $0$ {\color{gray}{\tiny $\pm 0$}} & $0$ {\color{gray}{\tiny $\pm 0$}} & $0$ {\color{gray}{\tiny $\pm 0$}} & $0$ {\color{gray}{\tiny $\pm 0$}} & $4$ {\color{gray}{\tiny $\pm 4$}} & $3$ {\color{gray}{\tiny $\pm 2$}} & $9$ {\color{gray}{\tiny $\pm 6$}} & $\mathbf{60}$ {\color{gray}{\tiny $\pm 2$}}\\
\texttt{humanoidmaze-med-st ($\mathbf{5}$)} & $2$ {\color{gray}{\tiny $\pm 1$}} & $33$ {\color{gray}{\tiny $\pm 2$}} & $22$ {\color{gray}{\tiny $\pm 8$}} & $1$ {\color{gray}{\tiny $\pm 0$}} & $1$ {\color{gray}{\tiny $\pm 1$}} & $53$ {\color{gray}{\tiny $\pm 8$}} & $19$ {\color{gray}{\tiny $\pm 1$}} & $38$ {\color{gray}{\tiny $\pm 5$}} & $\mathbf{60}$ {\color{gray}{\tiny $\pm 1$}} & $58$ {\color{gray}{\tiny $\pm 5$}} & $\mathbf{62}$ {\color{gray}{\tiny $\pm 2$}} \\
\texttt{humanoidmaze-large-st ($\mathbf{5}$)} & $1$ {\color{gray}{\tiny $\pm 0$}} & $2$ {\color{gray}{\tiny $\pm 1$}} & $2$ {\color{gray}{\tiny $\pm 1$}} & $1$ {\color{gray}{\tiny $\pm 0$}} & $0$ {\color{gray}{\tiny $\pm 0$}} & $0$ {\color{gray}{\tiny $\pm 0$}} & $0$ {\color{gray}{\tiny $\pm 0$}} & $2$ {\color{gray}{\tiny $\pm 0$}} & $\mathbf{11}$ {\color{gray}{\tiny $\pm 2$}} & $4$ {\color{gray}{\tiny $\pm 2$}} & $5$ {\color{gray}{\tiny $\pm 8$}} \\
\texttt{antsoccer-arena-st ($\mathbf{5}$)} & $1$ {\color{gray}{\tiny $\pm 0$}} & $8$ {\color{gray}{\tiny $\pm 2$}} & $0$ {\color{gray}{\tiny $\pm 0$}} & $12$ {\color{gray}{\tiny $\pm 4$}} & $1$ {\color{gray}{\tiny $\pm 0$}} & $2$ {\color{gray}{\tiny $\pm 4$}} & $12$ {\color{gray}{\tiny $\pm 0$}} & $16$ {\color{gray}{\tiny $\pm 1$}} & $33$ {\color{gray}{\tiny $\pm 6$}} & $60$ {\color{gray}{\tiny $\pm 2$}} & $\mathbf{65}$ {\color{gray}{\tiny $\pm 2$}}\\
\texttt{cube-single-st ($\mathbf{5}$)} & $5$ {\color{gray}{\tiny $\pm 1$}} & $83$ {\color{gray}{\tiny $\pm 3$}} & $91$ {\color{gray}{\tiny $\pm 2$}} & $\mathbf{95}$ {\color{gray}{\tiny $\pm 2$}} & $80$ {\color{gray}{\tiny $\pm 5$}} & $85$ {\color{gray}{\tiny $\pm 9$}} & $81$ {\color{gray}{\tiny $\pm 4$}} & $79$ {\color{gray}{\tiny $\pm 7$}} & $79$ {\color{gray}{\tiny $\pm 2$}} & $\mathbf{96}$ {\color{gray}{\tiny $\pm 1$}} & $\mathbf{94}$ {\color{gray}{\tiny $\pm 3$}} \\
\texttt{cube-double-st ($\mathbf{5}$)} & $2$ {\color{gray}{\tiny $\pm 1$}} & $7$ {\color{gray}{\tiny $\pm 1$}} & $12$ {\color{gray}{\tiny $\pm 1$}} & $15$ {\color{gray}{\tiny $\pm 6$}} & $2$ {\color{gray}{\tiny $\pm 1$}} & $6$ {\color{gray}{\tiny $\pm 2$}} & $5$ {\color{gray}{\tiny $\pm 2$}} & $15$ {\color{gray}{\tiny $\pm 3$}} & $14$ {\color{gray}{\tiny $\pm 3$}} & $\mathbf{29}$ {\color{gray}{\tiny $\pm 2$}} & $25$ {\color{gray}{\tiny $\pm 2$}} \\
\texttt{scene-st ($\mathbf{5}$)} & $5$ {\color{gray}{\tiny $\pm 1$}} & $28$ {\color{gray}{\tiny $\pm 1$}} & $41$ {\color{gray}{\tiny $\pm 3$}} & $46$ {\color{gray}{\tiny $\pm 3$}} & $20$ {\color{gray}{\tiny $\pm 1$}} & $40$ {\color{gray}{\tiny $\pm 7$}} & $30$ {\color{gray}{\tiny $\pm 3$}} & $45$ {\color{gray}{\tiny $\pm 5$}} & $30$ {\color{gray}{\tiny $\pm 3$}} & $56$ {\color{gray}{\tiny $\pm 2$}} & $\mathbf{74}$ {\color{gray}{\tiny $\pm 4$}} \\
\texttt{puzzle-3x3-st ($\mathbf{5}$)} & $2$ {\color{gray}{\tiny $\pm 0$}} & $9$ {\color{gray}{\tiny $\pm 1$}} & $21$ {\color{gray}{\tiny $\pm 1$}} & $10$ {\color{gray}{\tiny $\pm 2$}} & $18$ {\color{gray}{\tiny $\pm 1$}} & $19$ {\color{gray}{\tiny $\pm 0$}} & $6$ {\color{gray}{\tiny $\pm 2$}} & $14$ {\color{gray}{\tiny $\pm 4$}} & $19$ {\color{gray}{\tiny $\pm 1$}} & $30$ {\color{gray}{\tiny $\pm 1$}} & $\mathbf{35}$ {\color{gray}{\tiny $\pm 3$}} \\
\texttt{puzzle-4x4-st ($\mathbf{5}$)} & $0$ {\color{gray}{\tiny $\pm 0$}} & $7$ {\color{gray}{\tiny $\pm 1$}} & $14$ {\color{gray}{\tiny $\pm 1$}} & $\mathbf{29}$ {\color{gray}{\tiny $\pm 3$}} & $10$ {\color{gray}{\tiny $\pm 3$}} & $15$ {\color{gray}{\tiny $\pm 3$}} & $1$ {\color{gray}{\tiny $\pm 0$}} & $13$ {\color{gray}{\tiny $\pm 1$}} & $25$ {\color{gray}{\tiny $\pm 5$}} & $17$ {\color{gray}{\tiny $\pm 2$}} & $\mathbf{27}$ {\color{gray}{\tiny $\pm 3$}} \\
\addlinespace
\cmidrule(lr){2-12}
\textbf{\texttt{OGBench Overall}} & $3$ {\color{gray}{\tiny $\pm 0$}} & $23$ {\color{gray}{\tiny $\pm 1$}} & $31$ {\color{gray}{\tiny $\pm 1$}} & $23$ {\color{gray}{\tiny $\pm 1$}} & $14$ {\color{gray}{\tiny $\pm 1$}} & $25$ {\color{gray}{\tiny $\pm 2$}} & $16$ {\color{gray}{\tiny $\pm 1$}} & $29$ {\color{gray}{\tiny $\pm 1$}} & $30$ {\color{gray}{\tiny $\pm 1$}} & $44$ {\color{gray}{\tiny $\pm 1$}} & $\mathbf{54}$ {\color{gray}{\tiny $\pm 1$}} \\
\bottomrule
\end{tabular}
\end{threeparttable}
}
\end{table*}

\textbf{Training and Evaluation.} We train \ourMethod for 1M gradient steps on state-based OGBench tasks, and 500K steps on D4RL tasks, evaluating the agent every 100K steps. To prevent evaluation bias, we assess \ourMethod and all baselines using a fixed number of gradient steps rather than selecting the best performance across epochs. For OGBench, we adhere to the official evaluation protocol \citep{ogbench_park2025} and report the \textit{average success rate across the last three evaluation epochs}.
For D4RL, we follow standard practice and report \textit{performance at the last epoch} \citep{tarasov2023corl}. Performances are averaged over 8 seeds for all tasks.

\subsection{Main Results: Downstream Offline RL Performance}\label{sec:exp_main}
We report success rates on OGBench and normalized scores on D4RL, comparing \ourMethod against both unimodal baselines \citep{kostrikov2021iql, tarasov2023revisiting} and expressive multi-step generative baselines \citep{hansen2023idql, fql_park2025}. Baseline numbers are taken from prior work where available and reproduced with official implementations otherwise with full sourcing detailed in \aref{app:full_results}. Our goal is to establish that \ourMethod is competitive on complex continuous control without iterative inference.

\autoref{table:offline} shows that \ourMethod is competitive across the full suite, with the largest gains on the hardest tasks. It substantially outperforms all prior methods on long-horizon navigation (\texttt{antmaze-large-st}, \texttt{antmaze-giant-st}), \texttt{scene-st}, and \texttt{antsoccer-arena-st}, exceeds baselines on \texttt{humanoidmaze-medium-st} and \texttt{puzzle-3x3-st}, and matches the strongest baseline on \texttt{puzzle-4x4-st} and D4RL \texttt{antmaze}. Per-task results are reported in \aref{app:full_results}.

\subsection{Robustness Under Corrupted Offline Data}
\label{sec:intuition_corruption}

We next ask whether the same advantage as \autoref{sec:exp_main} appears when the offline dataset itself becomes less reliable. We use the default \texttt{cube-single} and \texttt{antmaze-large} tasks from OGBench and modify the official data collection rule by replacing a fraction \(p \in \{0.20, 0.30, 0.40\}\) of collected actions with uniformly random actions. Following the protocol used in the main experiments, we report success averaged over the \(800\)k, \(900\)k, and \(1\)M checkpoints. We focus on these default tasks because the compared methods are already competitive in the corresponding clean-data settings, so the corruption study isolates robustness rather than trivial clean-data weakness.

\begin{figure}[t]
    \centering
    \includegraphics[width=0.9\linewidth]{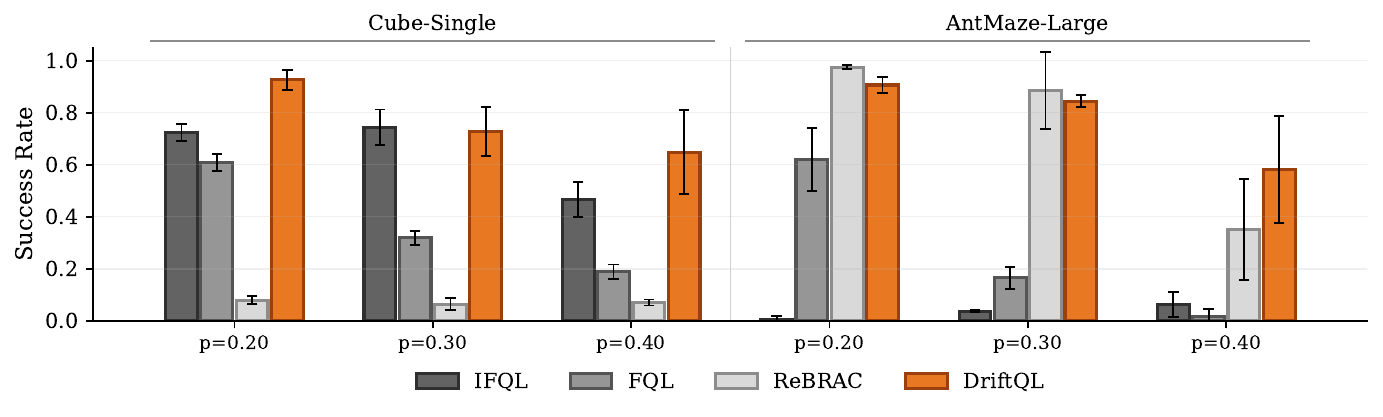}
    \caption{\textbf{Robustness under random-action corruption.} Success rate on the default \texttt{cube-single} and \texttt{antmaze-large} tasks as the random-action fraction increases from \(p=0.20\) to \(p=0.40\). Results are averaged over the \(800\)k, \(900\)k, and \(1\)M evaluation checkpoints and shown with 95\% confidence intervals. }
    \label{fig:intuition_corruption}
\end{figure}

\autoref{fig:intuition_corruption} shows that the same picture reappears under corrupted offline data. On Cube, \ourMethod is strongest at every corruption level. On AntMaze, ReBRAC is stronger at milder corruption, but \ourMethod degrades more gracefully and becomes the strongest method at the hardest setting, \(p=0.40\).

\subsection{Computational Efficiency and Inference Latency}\label{sec:exp_latency}
\begin{figure}[tb]
    \centering
    \begin{minipage}{0.55\textwidth}
        \centering
        \small
        \captionof{table}{Inference complexity of expressive offline RL methods. \textit{Sequential NFEs} is the number of network evaluations that cannot be parallelized, i.e.\ the main bottleneck for inference speed. IDQL and IFQL generate $N_c$ candidates in parallel, each taking $K$ sequential steps. \ourMethod has only a single forward pass with no distillation.}
        \label{tab:latency}
        \resizebox{\linewidth}{!}{%
        \begin{tabular}{lcc}
        \toprule
        \textbf{Method} & \textbf{Sequential NFEs} & \textbf{Distillation required} \\
        \midrule
        Diffusion-QL \citep{diffusionQL} & $K \times N_c$       & No  \\
        IDQL \citep{hansen2023idql}   & $K \times N_c$              & No  \\
        FQL  \citep{fql_park2025}     & 5--20 (ODE) + 1 (distilled) & Yes \\
        IFQL \citep{fql_park2025}     & $K \times N_c$              & No  \\
        \textbf{\ourMethod}           & \textbf{1 (feedforward)}    & \textbf{No} \\
        \bottomrule
        \end{tabular}%
        }
    \end{minipage}\hfill
    \begin{minipage}{0.4\textwidth}
        \centering
        \includegraphics[width=\linewidth]{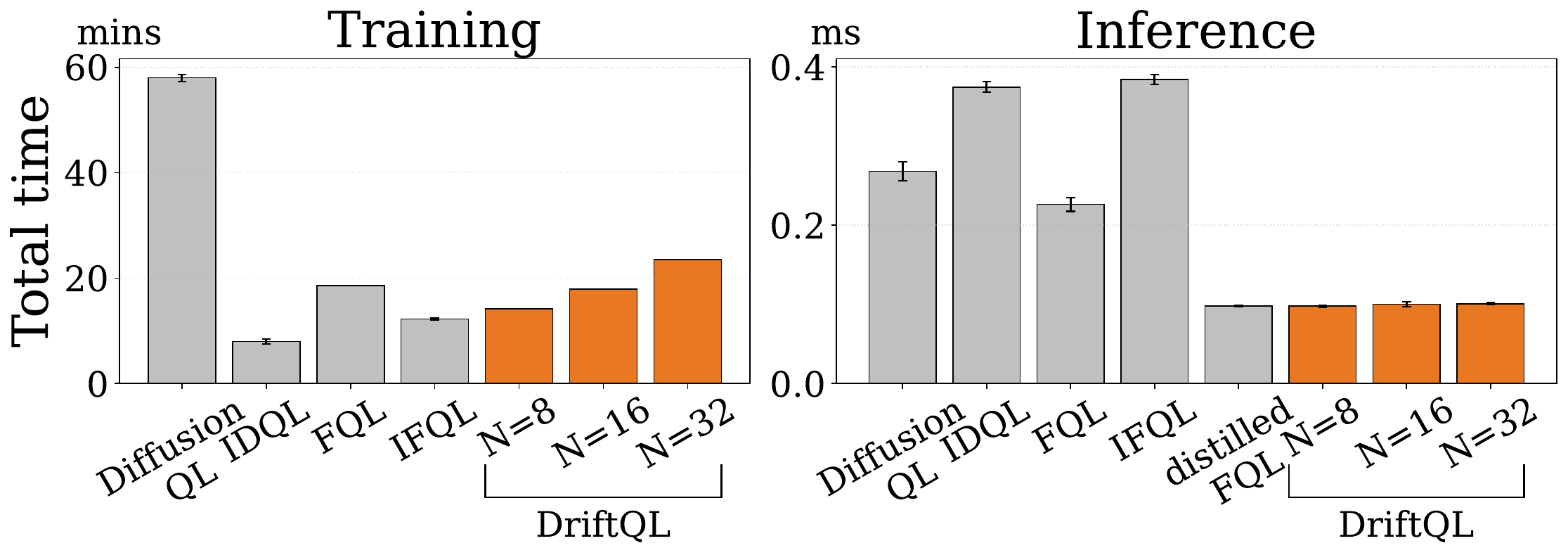}
        \caption{Total training time for 1M steps (left, minutes) and inference latency per step (right, ms), measured on default task from \texttt{antmaze-large-st}. All \ourMethod variants share the same single-pass inference cost. Training cost scales with $N$.}
        \label{fig:latency}
    \end{minipage}
\end{figure}

While diffusion \citep{ho2020ddpm_diffusion} and flow-matching models \citep{Lipman2022FlowMF} require multiple sequential network evaluations at inference, \ourMethod generates actions in a single forward pass. \cref{tab:latency} summarizes the per-method inference complexity, and \autoref{fig:latency} reports wall-clock measurements on the default \texttt{antmaze-large-st} task, averaged over 3 seeds on a single NVIDIA RTX 4090. To ensure a fair comparison, all methods use their tuned hyperparameters and share identical actor and critic network sizes.

\textbf{Inference.} \ourMethod matches distilled FQL without requiring a separate distillation stage, and runs roughly $2{\times}$ faster than FQL, $3{\times}$ faster than Diffusion-QL, and $4{\times}$ faster than IDQL and IFQL. This advantage stems from the single-pass design: diffusion- and flow-based methods incur sequential ODE or denoising steps, while \ourMethod transports the distribution during training and emits actions directly at test time. Since inference uses one feedforward pass regardless of $N$, all \ourMethod variants share identical latency.

\textbf{Training.} Training cost grows with $N$ due to the $O(N^2)$ pairwise kernel, but the absolute overhead remains modest. At $N=8$, \ourMethod trains faster than FQL, at $N=16$, it matches FQL, and at $N=32$, it is only marginally slower. Diffusion-QL is substantially slower than all flow- and drift-based methods due to its longer denoising chain at training time. Larger $N$ thus trades minor additional compute for greater sample diversity in the repulsion field.
\hc{why should DriftQL converge faster? there is no theoretical justification on this}

\subsection{Ablation Study}
\label{sec:exp_ablations}

\begin{figure}[t]
    \centering
    \includegraphics[width=\linewidth]{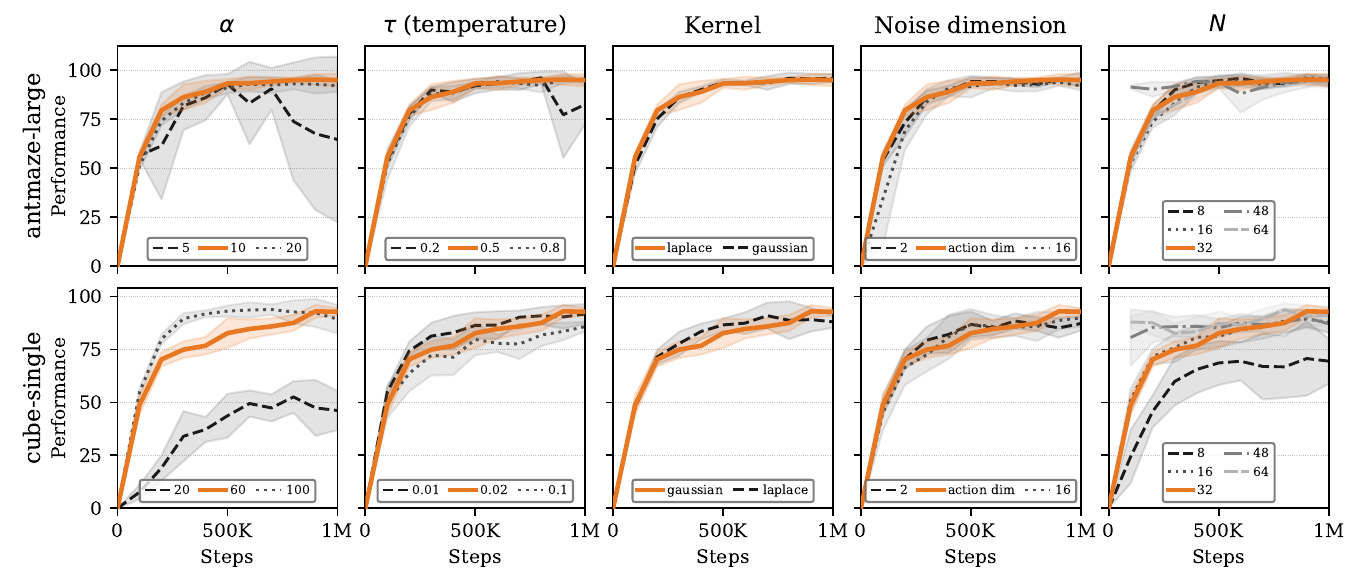}
    \caption{\textbf{Ablation Study of \ourMethod Hyperparameters.} Success rates across training steps for default tasks in \texttt{antmaze-large-navigate} (top) and \texttt{cube-single-play} (bottom), averaged over 3 seeds with $\pm1$ standard deviation shaded. From left to right: the behavioral regularization trade-off ($\alpha$), the drift kernel temperature ($\tau$), the kernel type (Laplace vs.\ Gaussian), the noise dimension ($z$ size), and the number of generated action samples per state ($N$). The default optimal setting for each environment is denoted by a \textcolor{myorange}{solid orange line}, while variations are plotted with grey lines.}
    \label{fig:ablations}
\end{figure}

We isolate the impact of \ourMethod's hyperparameters on two representative default environments that exercise different aspects of the drift field: \texttt{antmaze-large-navigate} (sparse-reward navigation, 29-dimensional observations and 8-dimensional actions) and \texttt{cube-single-play} (robotic cube manipulation, 28-dimensional observations and 5-dimensional actions). All ablations are averaged over 3 seeds, with one variable changed at a time relative to the per-environment defaults specified in \aref{app:hyperparams}. Training curves for each ablation are shown in \autoref{fig:ablations}.

\textbf{Trade-off parameter $\alpha$}:
The scalar $\alpha$ governs the balance between the drift behavioral regularizer and the Q-maximization objective (\autoref{eq:actor_loss}), and it is the most important hyperparameter of \ourMethod, consistent with the broader offline RL literature \citep{fujimoto2021td3bc, tarasov2023revisiting, fql_park2025}. As shown in \autoref{fig:ablations} (\nth{1} col), the sensitivity of performance to $\alpha$ differs markedly across environments: on \texttt{antmaze-large}, undersized $\alpha$ leads to a late-training collapse, while a slightly oversized $\alpha$ stays close to the default. Yet, on \texttt{cube-single}, the degrades are more noticeable at the extremes. Intuitively, when $\alpha$ is too small the drift constraint provides insufficient support in low-density regions of the dataset, whereas an overly large $\alpha$ suppresses the Q-gradient and prevents improvement beyond the behavioral policy. We therefore recommend tuning $\alpha$ per environment, following the same protocol as \citet{fql_park2025}: a coarse sweep over one order of magnitude is generally sufficient to identify a well-performing setting.

\textbf{Kernel temperature $\tau$}:
The temperature $\tau$ controls the sharpness of the kernel-weighted repulsion (\autoref{eq:kernel}): small $\tau$ concentrates repulsive force on the nearest neighbors, while large $\tau$ spreads it more uniformly across all generated samples. \autoref{fig:ablations} (\nth{2} col) shows that \ourMethod is broadly robust to $\tau$ across both environments, with performance remaining stable over a wide range. The curves separate only at the extremes, where very small $\tau$ can destabilize training by concentrating all repulsive mass on a single neighbor, and very large $\tau$ effectively removes the distance-sensitivity of the kernel. In practice, the default $\tau$ can be used without tuning.

\textbf{Kernel choice}:
\ourMethod uses a Gaussian kernel for the repulsion weights by default, motivated by its exact connection to the reverse-Fisher divergence on smoothed distributions \citep{lai2026unified}. \autoref{fig:ablations} (\nth{3} col) compares the Gaussian kernel against the Laplace alternative used in the original drifting-model implementation \citep{driftmodels}. On \texttt{antmaze-large}, the default and the softer setting are essentially indistinguishable, while the sharpest setting trails behind, with the gap appearing only late in training. On \texttt{cube-single}, although both kernels perform similarly throughout training, the Gaussian kernel provides an edge late in training.

\textbf{Noise dimension}:
The actor $f_\theta(s, \epsilon)$ takes a noise vector $\epsilon \sim \mathcal{N}(0, I)$ whose dimension defaults to the action dimension. \autoref{fig:ablations} (\nth{4} col) shows that this choice is essentially free across a wide range. On \texttt{antmaze-large}, the smaller, default, and larger noise dimensions all converge to comparable performance, with only the larger setting showing slightly slower early progress before catching up. On \texttt{cube-single}, the picture is similar, with all settings performing similarly, except in the last few steps in training that they deviate slightly. We use the action-dimension default for simplicity.

\textbf{Number of generated samples $N_\text{gen}$}:
The repulsive component of the drift field is computed over $N$ generated samples per state. Because each step incurs an $O(N^2)$ pairwise kernel cost, $N$ trades compute against the fidelity of the empirical negative distribution. \autoref{fig:ablations} (\nth{5} col) shows how performance varies with $N$ across both environments. On \texttt{antmaze-large}, all values for $N$ perform similarly, while on \texttt{cube-single}, the smaller $N$ struggles to keep up with the default and larger values, which are comparable. On these two ablation environments, $N\ge16$ essentially converges to the same performance, but on a broader sweep across the full benchmark we observed cases where $N=16$ underperformed while $N=32$ remained stable. Moreover, $N\ge32$ leads to computational overhead while providing minimal improvements. Given these observations, we adopt $N=32$ as the default to ensure robustness across environments without retuning per task.

\section{Conclusion}\label{sec:conclusion}

We introduced \ourMethod, an offline RL learner based on a one-step generative actor and a state-conditioned drifting objective. The method uses a drift regularizer to keep generated actions near dataset support and a critic objective to drive policy improvement. Unlike diffusion and flow policies that require iterative sampling, solvers, or distillation for fast inference, \ourMethod trains a single stochastic actor that produces actions with one forward pass. Empirically, \ourMethod performs competitively across D4RL and OGBench, with especially strong gains on difficult OGBench navigation and manipulation tasks. It also remains robust under random-action data corruption and retains the inference efficiency of deterministic one-step policies. These results suggest that drifting provides a useful middle ground for offline RL: it gives the actor a stochastic, distributional training signal while avoiding the test-time cost of multi-step generative policies.

Several questions remain open. We provide a detailed discussion of limitations in \autoref{app:limitations}, including the training-time cost of pairwise generated-action interactions, and the assumption of continuous box-bounded action spaces. It would also be useful to study more adaptive drift estimators, including learned or state-dependent kernels and cheaper approximations to the generated-action repulsion term. Finally, this work focuses on purely offline training with low-dimensional state spaces. Extending \ourMethod to high-dimensional observations such as images, as well as to offline-to-online fine-tuning and online RL where fast action sampling and continued policy improvement are both important, are natural next steps.

\clearpage
\bibliographystyle{plainnat}
\bibliography{references}

\input{appendix}


\end{document}

%% file: math_commands.tex
\usepackage{amsmath,amsfonts,bm}

\usepackage{xspace}

\def\eqref#1{equation~\ref{#1}}

\def\1{\bm{1}}

\DeclareMathAlphabet{\mathsfit}{\encodingdefault}{\sfdefault}{m}{sl}
\SetMathAlphabet{\mathsfit}{bold}{\encodingdefault}{\sfdefault}{bx}{n}



%% file: appendix.tex
\clearpage
\appendix
\doparttoc
\faketableofcontents

\setcounter{parttocdepth}{1}

\renewcommand{\thepart}{}

\part{Appendix}\label{appendix}
\parttoc

\clearpage

\section{Full Experimental Results}
\label{app:full_results}

\autoref{table:offline_full} reports the complete, per-task performance of \ourMethod and all baselines across the full OGBench and D4RL benchmark tasks. The main text (\autoref{sec:exp_main}) condenses these into environment-level averages to summarize overall trends. This table provides the unaggregated breakdown to enable full reproducibility and to facilitate fine-grained comparisons for future work.

\textbf{Table structure.} Rows correspond to individual tasks, grouped by environment. The \texttt{(*)} marker denotes the default task used for hyperparameter tuning within each environment in OGBench. All other tasks in the group use the same hyperparameters, following the evaluation protocol of \citet{fql_park2025}. Columns are organized by policy class: Gaussian policies (BC, IQL, ReBRAC), diffusion policies (IDQL, SRPO, CAC), flow policies (FAWAC, FBRAC, IFQL, FQL), and \ourMethod. Values are averaged over 8 seeds. Standard deviations are reported after ``$\pm$''. Values at or above 95\% of the best performance in each row are \textbf{bolded}, following \citet{ogbench_park2025}.

\textbf{Sources of baseline numbers.} For OGBench tasks and the D4RL Antmaze and Adroit suites, we report baseline numbers directly from \citet{fql_park2025}. For D4RL Locomotion, baseline numbers are taken from the original papers: BC, IQL, and ReBRAC from \citet{tarasov2023corl}, IDQL from \citet{hansen2023idql}, and SRPO from \citet{chen2024score}. For IFQL and FQL on D4RL Locomotion, no published numbers were available for the exact task variants we evaluate, so we ran both methods ourselves using the official author implementations and tuned hyperparameters following the protocol described by \citet{fql_park2025}. For methods denoted with $-$, due to computational constraints we were unable to obtain reliable results within the allotted budget, and no published numbers were available for those task variants to the best of our knowledge.

\begin{table*}[t]
\caption{
\footnotesize
\textbf{Full offline RL results.}
We present the full results on the $77$ OGBench and D4RL tasks. \texttt{(*)} indicates the default task in each environment.
The results are averaged over $8$ seeds ($4$ seeds for pixel-based tasks) unless otherwise mentioned.
}
\label{table:offline_full}
\centering
\resizebox{\textwidth}{!}
{
\begin{threeparttable}
\begin{tabular}{lccccccccccc}
\toprule
\multicolumn{1}{c}{} & \multicolumn{3}{c}{\texttt{Gaussian Policies}} & \multicolumn{3}{c}{\texttt{Diffusion Policies}} & \multicolumn{4}{c}{\texttt{Flow Policies}} & \multicolumn{1}{c}{\texttt{Drift}} \\
\cmidrule(lr){2-4} \cmidrule(lr){5-7} \cmidrule(lr){8-11} \cmidrule(lr){12-12}
\texttt{Task} & \texttt{BC} & \texttt{IQL} & \texttt{ReBRAC} & \texttt{IDQL} & \texttt{SRPO} & \texttt{CAC} & \texttt{FAWAC} & \texttt{FBRAC} & \texttt{IFQL} & \texttt{FQL} & \texttt{\color{myorange}\ourMethod} \\
\midrule
\texttt{antmaze-large-navigate-singletask-task1-v0 (*)} & $0$ {\tiny $\pm 0$} & $48$ {\tiny $\pm 9$} & $\mathbf{91}$ {\tiny $\pm 10$} & $0$ {\tiny $\pm 0$} & $0$ {\tiny $\pm 0$} & $42$ {\tiny $\pm 7$} & $1$ {\tiny $\pm 1$} & $70$ {\tiny $\pm 20$} & $24$ {\tiny $\pm 17$} & $80$ {\tiny $\pm 8$} & $\mathbf{95}$ {\tiny $\pm 2$} \\
\texttt{antmaze-large-navigate-singletask-task2-v0} & $6$ {\tiny $\pm 3$} & $42$ {\tiny $\pm 6$} & $\mathbf{88}$ {\tiny $\pm 4$} & $14$ {\tiny $\pm 8$} & $4$ {\tiny $\pm 4$} & $1$ {\tiny $\pm 1$} & $0$ {\tiny $\pm 1$} & $35$ {\tiny $\pm 12$} & $8$ {\tiny $\pm 3$} & $57$ {\tiny $\pm 10$} & $\mathbf{85}$ {\tiny $\pm 9$} \\
\texttt{antmaze-large-navigate-singletask-task3-v0} & $29$ {\tiny $\pm 5$} & $72$ {\tiny $\pm 7$} & $51$ {\tiny $\pm 18$} & $26$ {\tiny $\pm 8$} & $3$ {\tiny $\pm 2$} & $49$ {\tiny $\pm 10$} & $12$ {\tiny $\pm 4$} & $83$ {\tiny $\pm 15$} & $52$ {\tiny $\pm 17$} & $\mathbf{93}$ {\tiny $\pm 3$} & $\mathbf{97}$ {\tiny $\pm 1$} \\
\texttt{antmaze-large-navigate-singletask-task4-v0} & $8$ {\tiny $\pm 3$} & $51$ {\tiny $\pm 9$} & $84$ {\tiny $\pm 7$} & $62$ {\tiny $\pm 25$} & $45$ {\tiny $\pm 19$} & $17$ {\tiny $\pm 6$} & $10$ {\tiny $\pm 3$} & $37$ {\tiny $\pm 18$} & $18$ {\tiny $\pm 8$} & $80$ {\tiny $\pm 4$} & $\mathbf{91}$ {\tiny $\pm 1$} \\
\texttt{antmaze-large-navigate-singletask-task5-v0} & $10$ {\tiny $\pm 3$} & $54$ {\tiny $\pm 22$} & $\mathbf{90}$ {\tiny $\pm 2$} & $2$ {\tiny $\pm 2$} & $1$ {\tiny $\pm 1$} & $55$ {\tiny $\pm 6$} & $9$ {\tiny $\pm 5$} & $76$ {\tiny $\pm 8$} & $38$ {\tiny $\pm 18$} & $83$ {\tiny $\pm 4$} & $\mathbf{92}$ {\tiny $\pm 1$} \\
\midrule
\texttt{antmaze-giant-navigate-singletask-task1-v0 (*)} & $0$ {\tiny $\pm 0$} & $0$ {\tiny $\pm 0$} & $27$ {\tiny $\pm 22$} & $0$ {\tiny $\pm 0$} & $0$ {\tiny $\pm 0$} & $0$ {\tiny $\pm 0$} & $0$ {\tiny $\pm 0$} & $0$ {\tiny $\pm 1$} & $0$ {\tiny $\pm 0$} & $4$ {\tiny $\pm 5$} & $\mathbf{32}$ {\tiny $\pm 2$} \\
\texttt{antmaze-giant-navigate-singletask-task2-v0} & $0$ {\tiny $\pm 0$} & $1$ {\tiny $\pm 1$} & $16$ {\tiny $\pm 17$} & $0$ {\tiny $\pm 0$} & $0$ {\tiny $\pm 0$} & $0$ {\tiny $\pm 0$} & $0$ {\tiny $\pm 0$} & $4$ {\tiny $\pm 7$} & $0$ {\tiny $\pm 0$} & $9$ {\tiny $\pm 7$} & $\mathbf{79}$ {\tiny $\pm 1$} \\
\texttt{antmaze-giant-navigate-singletask-task3-v0} & $0$ {\tiny $\pm 0$} & $0$ {\tiny $\pm 0$} & $34$ {\tiny $\pm 22$} & $0$ {\tiny $\pm 0$} & $0$ {\tiny $\pm 0$} & $0$ {\tiny $\pm 0$} & $0$ {\tiny $\pm 0$} & $0$ {\tiny $\pm 0$} & $0$ {\tiny $\pm 0$} & $0$ {\tiny $\pm 1$} & $\mathbf{43}$ {\tiny $\pm 2$} \\
\texttt{antmaze-giant-navigate-singletask-task4-v0} & $0$ {\tiny $\pm 0$} & $0$ {\tiny $\pm 0$} & $5$ {\tiny $\pm 12$} & $0$ {\tiny $\pm 0$} & $0$ {\tiny $\pm 0$} & $0$ {\tiny $\pm 0$} & $0$ {\tiny $\pm 0$} & $9$ {\tiny $\pm 4$} & $0$ {\tiny $\pm 0$} & $14$ {\tiny $\pm 23$} & $\mathbf{64}$ {\tiny $\pm 3$} \\
\texttt{antmaze-giant-navigate-singletask-task5-v0} & $1$ {\tiny $\pm 1$} & $19$ {\tiny $\pm 7$} & $49$ {\tiny $\pm 22$} & $0$ {\tiny $\pm 1$} & $0$ {\tiny $\pm 0$} & $0$ {\tiny $\pm 0$} & $0$ {\tiny $\pm 0$} & $6$ {\tiny $\pm 10$} & $13$ {\tiny $\pm 9$} & $16$ {\tiny $\pm 28$} & $\mathbf{85}$ {\tiny $\pm 5$} \\
\midrule
\texttt{humanoidmaze-medium-navigate-singletask-task1-v0 (*)} & $1$ {\tiny $\pm 0$} & $32$ {\tiny $\pm 7$} & $16$ {\tiny $\pm 9$} & $1$ {\tiny $\pm 1$} & $0$ {\tiny $\pm 0$} & $38$ {\tiny $\pm 19$} & $6$ {\tiny $\pm 2$} & $25$ {\tiny $\pm 8$} & $\mathbf{69}$ {\tiny $\pm 19$} & $19$ {\tiny $\pm 12$} & $28$ {\tiny $\pm 2$} \\
\texttt{humanoidmaze-medium-navigate-singletask-task2-v0} & $1$ {\tiny $\pm 0$} & $41$ {\tiny $\pm 9$} & $18$ {\tiny $\pm 16$} & $1$ {\tiny $\pm 1$} & $1$ {\tiny $\pm 1$} & $47$ {\tiny $\pm 35$} & $40$ {\tiny $\pm 2$} & $76$ {\tiny $\pm 10$} & $85$ {\tiny $\pm 11$} & $\mathbf{94}$ {\tiny $\pm 3$} & $87$ {\tiny $\pm 2$} \\
\texttt{humanoidmaze-medium-navigate-singletask-task3-v0} & $6$ {\tiny $\pm 2$} & $25$ {\tiny $\pm 5$} & $36$ {\tiny $\pm 13$} & $0$ {\tiny $\pm 1$} & $2$ {\tiny $\pm 1$} & $\mathbf{83}$ {\tiny $\pm 18$} & $19$ {\tiny $\pm 2$} & $27$ {\tiny $\pm 11$} & $49$ {\tiny $\pm 49$} & $74$ {\tiny $\pm 18$} & $56$ {\tiny $\pm 3$} \\
\texttt{humanoidmaze-medium-navigate-singletask-task4-v0} & $0$ {\tiny $\pm 0$} & $0$ {\tiny $\pm 1$} & $15$ {\tiny $\pm 16$} & $1$ {\tiny $\pm 1$} & $1$ {\tiny $\pm 1$} & $5$ {\tiny $\pm 4$} & $1$ {\tiny $\pm 1$} & $1$ {\tiny $\pm 2$} & $1$ {\tiny $\pm 1$} & $3$ {\tiny $\pm 4$} & $\mathbf{39}$ {\tiny $\pm 3$} \\
\texttt{humanoidmaze-medium-navigate-singletask-task5-v0} & $2$ {\tiny $\pm 1$} & $66$ {\tiny $\pm 4$} & $24$ {\tiny $\pm 20$} & $1$ {\tiny $\pm 1$} & $3$ {\tiny $\pm 3$} & $91$ {\tiny $\pm 5$} & $31$ {\tiny $\pm 7$} & $63$ {\tiny $\pm 9$} & $\mathbf{98}$ {\tiny $\pm 2$} & $\mathbf{97}$ {\tiny $\pm 2$} & $\mathbf{99}$ {\tiny $\pm 2$} \\
\midrule
\texttt{humanoidmaze-large-navigate-singletask-task1-v0 (*)} & $0$ {\tiny $\pm 0$} & $3$ {\tiny $\pm 1$} & $2$ {\tiny $\pm 1$} & $0$ {\tiny $\pm 0$} & $0$ {\tiny $\pm 0$} & $1$ {\tiny $\pm 1$} & $0$ {\tiny $\pm 0$} & $0$ {\tiny $\pm 1$} & $6$ {\tiny $\pm 2$} & $\mathbf{7}$ {\tiny $\pm 6$} & $2$ {\tiny $\pm 1$} \\
\texttt{humanoidmaze-large-navigate-singletask-task2-v0} & $\mathbf{0}$ {\tiny $\pm 0$} & $\mathbf{0}$ {\tiny $\pm 0$} & $\mathbf{0}$ {\tiny $\pm 0$} & $\mathbf{0}$ {\tiny $\pm 0$} & $\mathbf{0}$ {\tiny $\pm 0$} & $\mathbf{0}$ {\tiny $\pm 0$} & $\mathbf{0}$ {\tiny $\pm 0$} & $\mathbf{0}$ {\tiny $\pm 0$} & $\mathbf{0}$ {\tiny $\pm 0$} & $\mathbf{0}$ {\tiny $\pm 0$} & $\mathbf{0}$ {\tiny $\pm 0$} \\
\texttt{humanoidmaze-large-navigate-singletask-task3-v0} & $1$ {\tiny $\pm 1$} & $7$ {\tiny $\pm 3$} & $8$ {\tiny $\pm 4$} & $3$ {\tiny $\pm 1$} & $1$ {\tiny $\pm 1$} & $2$ {\tiny $\pm 3$} & $1$ {\tiny $\pm 1$} & $10$ {\tiny $\pm 2$} & $\mathbf{48}$ {\tiny $\pm 10$} & $11$ {\tiny $\pm 7$} & $23$ {\tiny $\pm 1$} \\
\texttt{humanoidmaze-large-navigate-singletask-task4-v0} & $1$ {\tiny $\pm 0$} & $1$ {\tiny $\pm 0$} & $1$ {\tiny $\pm 1$} & $0$ {\tiny $\pm 0$} & $0$ {\tiny $\pm 0$} & $0$ {\tiny $\pm 1$} & $0$ {\tiny $\pm 0$} & $0$ {\tiny $\pm 0$} & $1$ {\tiny $\pm 1$} & $\mathbf{2}$ {\tiny $\pm 3$} & $1$ {\tiny $\pm 1$} \\
\texttt{humanoidmaze-large-navigate-singletask-task5-v0} & $0$ {\tiny $\pm 1$} & $1$ {\tiny $\pm 1$} & $\mathbf{2}$ {\tiny $\pm 2$} & $0$ {\tiny $\pm 0$} & $0$ {\tiny $\pm 0$} & $0$ {\tiny $\pm 0$} & $0$ {\tiny $\pm 0$} & $1$ {\tiny $\pm 1$} & $0$ {\tiny $\pm 0$} & $1$ {\tiny $\pm 3$} & $1$ {\tiny $\pm 0$} \\
\midrule
\texttt{antsoccer-arena-navigate-singletask-task1-v0} & $2$ {\tiny $\pm 1$} & $14$ {\tiny $\pm 5$} & $0$ {\tiny $\pm 0$} & $44$ {\tiny $\pm 12$} & $2$ {\tiny $\pm 1$} & $1$ {\tiny $\pm 3$} & $22$ {\tiny $\pm 2$} & $17$ {\tiny $\pm 3$} & $61$ {\tiny $\pm 25$} & $\mathbf{77}$ {\tiny $\pm 4$} & $\mathbf{79}$ {\tiny $\pm 4$} \\
\texttt{antsoccer-arena-navigate-singletask-task2-v0} & $2$ {\tiny $\pm 2$} & $17$ {\tiny $\pm 7$} & $0$ {\tiny $\pm 1$} & $15$ {\tiny $\pm 12$} & $3$ {\tiny $\pm 1$} & $0$ {\tiny $\pm 0$} & $8$ {\tiny $\pm 1$} & $8$ {\tiny $\pm 2$} & $75$ {\tiny $\pm 3$} & $\mathbf{88}$ {\tiny $\pm 3$} & $\mathbf{91}$ {\tiny $\pm 2$} \\
\texttt{antsoccer-arena-navigate-singletask-task3-v0} & $0$ {\tiny $\pm 0$} & $6$ {\tiny $\pm 4$} & $0$ {\tiny $\pm 0$} & $0$ {\tiny $\pm 0$} & $0$ {\tiny $\pm 0$} & $8$ {\tiny $\pm 19$} & $11$ {\tiny $\pm 5$} & $16$ {\tiny $\pm 3$} & $14$ {\tiny $\pm 22$} & $\mathbf{61}$ {\tiny $\pm 6$} & $\mathbf{60}$ {\tiny $\pm 3$} \\
\texttt{antsoccer-arena-navigate-singletask-task4-v0 (*)} & $1$ {\tiny $\pm 0$} & $3$ {\tiny $\pm 2$} & $0$ {\tiny $\pm 0$} & $0$ {\tiny $\pm 1$} & $0$ {\tiny $\pm 0$} & $0$ {\tiny $\pm 0$} & $12$ {\tiny $\pm 3$} & $24$ {\tiny $\pm 4$} & $16$ {\tiny $\pm 9$} & $39$ {\tiny $\pm 6$} & $\mathbf{48}$ {\tiny $\pm 5$} \\
\texttt{antsoccer-arena-navigate-singletask-task5-v0} & $0$ {\tiny $\pm 0$} & $2$ {\tiny $\pm 2$} & $0$ {\tiny $\pm 0$} & $0$ {\tiny $\pm 0$} & $0$ {\tiny $\pm 0$} & $0$ {\tiny $\pm 0$} & $9$ {\tiny $\pm 2$} & $15$ {\tiny $\pm 4$} & $0$ {\tiny $\pm 1$} & $36$ {\tiny $\pm 9$} & $\mathbf{48}$ {\tiny $\pm 9$} \\
\midrule
\texttt{cube-single-play-singletask-task1-v0} & $10$ {\tiny $\pm 5$} & $88$ {\tiny $\pm 3$} & $89$ {\tiny $\pm 5$} & $\mathbf{95}$ {\tiny $\pm 2$} & $89$ {\tiny $\pm 7$} & $77$ {\tiny $\pm 28$} & $81$ {\tiny $\pm 9$} & $73$ {\tiny $\pm 33$} & $79$ {\tiny $\pm 4$} & $\mathbf{97}$ {\tiny $\pm 2$} & $\mathbf{94}$ {\tiny $\pm 3$} \\
\texttt{cube-single-play-singletask-task2-v0 (*)} & $3$ {\tiny $\pm 1$} & $85$ {\tiny $\pm 8$} & $92$ {\tiny $\pm 4$} & $\mathbf{96}$ {\tiny $\pm 2$} & $82$ {\tiny $\pm 16$} & $80$ {\tiny $\pm 30$} & $81$ {\tiny $\pm 9$} & $83$ {\tiny $\pm 13$} & $73$ {\tiny $\pm 3$} & $\mathbf{97}$ {\tiny $\pm 2$} & $\mathbf{93}$ {\tiny $\pm 2$} \\
\texttt{cube-single-play-singletask-task3-v0} & $9$ {\tiny $\pm 3$} & $91$ {\tiny $\pm 5$} & $93$ {\tiny $\pm 3$} & $\mathbf{99}$ {\tiny $\pm 1$} & $\mathbf{96}$ {\tiny $\pm 2$} & $\mathbf{98}$ {\tiny $\pm 1$} & $87$ {\tiny $\pm 4$} & $82$ {\tiny $\pm 12$} & $88$ {\tiny $\pm 4$} & $\mathbf{98}$ {\tiny $\pm 2$} & $\mathbf{95}$ {\tiny $\pm 2$} \\
\texttt{cube-single-play-singletask-task4-v0} & $2$ {\tiny $\pm 1$} & $73$ {\tiny $\pm 6$} & $\mathbf{92}$ {\tiny $\pm 3$} & $\mathbf{93}$ {\tiny $\pm 4$} & $70$ {\tiny $\pm 18$} & $\mathbf{91}$ {\tiny $\pm 2$} & $79$ {\tiny $\pm 6$} & $79$ {\tiny $\pm 20$} & $79$ {\tiny $\pm 6$} & $\mathbf{94}$ {\tiny $\pm 3$} & $\mathbf{92}$ {\tiny $\pm 3$} \\
\texttt{cube-single-play-singletask-task5-v0} & $3$ {\tiny $\pm 3$} & $78$ {\tiny $\pm 9$} & $87$ {\tiny $\pm 8$} & $\mathbf{90}$ {\tiny $\pm 6$} & $61$ {\tiny $\pm 12$} & $80$ {\tiny $\pm 20$} & $78$ {\tiny $\pm 10$} & $76$ {\tiny $\pm 33$} & $77$ {\tiny $\pm 7$} & $\mathbf{93}$ {\tiny $\pm 3$} & $\mathbf{90}$ {\tiny $\pm 3$} \\
\midrule
\texttt{cube-double-play-singletask-task1-v0} & $8$ {\tiny $\pm 3$} & $27$ {\tiny $\pm 5$} & $45$ {\tiny $\pm 6$} & $39$ {\tiny $\pm 19$} & $7$ {\tiny $\pm 6$} & $21$ {\tiny $\pm 8$} & $21$ {\tiny $\pm 7$} & $47$ {\tiny $\pm 11$} & $35$ {\tiny $\pm 9$} & $\mathbf{61}$ {\tiny $\pm 9$} & $49$ {\tiny $\pm 1$} \\
\texttt{cube-double-play-singletask-task2-v0 (*)} & $0$ {\tiny $\pm 0$} & $1$ {\tiny $\pm 1$} & $7$ {\tiny $\pm 3$} & $16$ {\tiny $\pm 10$} & $0$ {\tiny $\pm 0$} & $2$ {\tiny $\pm 2$} & $2$ {\tiny $\pm 1$} & $22$ {\tiny $\pm 12$} & $9$ {\tiny $\pm 5$} & $\mathbf{36}$ {\tiny $\pm 6$} & $23$ {\tiny $\pm 5$} \\
\texttt{cube-double-play-singletask-task3-v0} & $0$ {\tiny $\pm 0$} & $0$ {\tiny $\pm 0$} & $4$ {\tiny $\pm 1$} & $17$ {\tiny $\pm 8$} & $0$ {\tiny $\pm 1$} & $3$ {\tiny $\pm 1$} & $1$ {\tiny $\pm 1$} & $4$ {\tiny $\pm 2$} & $8$ {\tiny $\pm 5$} & $\mathbf{22}$ {\tiny $\pm 5$} & $9$ {\tiny $\pm 3$} \\
\texttt{cube-double-play-singletask-task4-v0} & $0$ {\tiny $\pm 0$} & $0$ {\tiny $\pm 0$} & $1$ {\tiny $\pm 1$} & $0$ {\tiny $\pm 1$} & $0$ {\tiny $\pm 0$} & $0$ {\tiny $\pm 1$} & $0$ {\tiny $\pm 0$} & $0$ {\tiny $\pm 1$} & $1$ {\tiny $\pm 1$} & $\mathbf{5}$ {\tiny $\pm 2$} & $3$ {\tiny $\pm 1$} \\
\texttt{cube-double-play-singletask-task5-v0} & $0$ {\tiny $\pm 0$} & $4$ {\tiny $\pm 3$} & $4$ {\tiny $\pm 2$} & $1$ {\tiny $\pm 1$} & $0$ {\tiny $\pm 0$} & $3$ {\tiny $\pm 2$} & $2$ {\tiny $\pm 1$} & $2$ {\tiny $\pm 2$} & $17$ {\tiny $\pm 6$} & $19$ {\tiny $\pm 10$} & $\mathbf{43}$ {\tiny $\pm 2$} \\
\midrule
\texttt{scene-play-singletask-task1-v0} & $19$ {\tiny $\pm 6$} & $94$ {\tiny $\pm 3$} & $\mathbf{95}$ {\tiny $\pm 2$} & $\mathbf{100}$ {\tiny $\pm 0$} & $94$ {\tiny $\pm 4$} & $\mathbf{100}$ {\tiny $\pm 1$} & $87$ {\tiny $\pm 8$} & $\mathbf{96}$ {\tiny $\pm 8$} & $\mathbf{98}$ {\tiny $\pm 3$} & $\mathbf{100}$ {\tiny $\pm 0$} & $\mathbf{100}$ {\tiny $\pm 0$} \\
\texttt{scene-play-singletask-task2-v0 (*)} & $1$ {\tiny $\pm 1$} & $12$ {\tiny $\pm 3$} & $50$ {\tiny $\pm 13$} & $33$ {\tiny $\pm 14$} & $2$ {\tiny $\pm 2$} & $50$ {\tiny $\pm 40$} & $18$ {\tiny $\pm 8$} & $46$ {\tiny $\pm 10$} & $0$ {\tiny $\pm 0$} & $76$ {\tiny $\pm 9$} & $\mathbf{89}$ {\tiny $\pm 3$} \\
\texttt{scene-play-singletask-task3-v0} & $1$ {\tiny $\pm 1$} & $32$ {\tiny $\pm 7$} & $55$ {\tiny $\pm 16$} & $\mathbf{94}$ {\tiny $\pm 4$} & $4$ {\tiny $\pm 4$} & $49$ {\tiny $\pm 16$} & $38$ {\tiny $\pm 9$} & $78$ {\tiny $\pm 14$} & $54$ {\tiny $\pm 19$} & $\mathbf{98}$ {\tiny $\pm 1$} & $\mathbf{93}$ {\tiny $\pm 3$} \\
\texttt{scene-play-singletask-task4-v0} & $2$ {\tiny $\pm 2$} & $0$ {\tiny $\pm 1$} & $3$ {\tiny $\pm 3$} & $4$ {\tiny $\pm 3$} & $0$ {\tiny $\pm 0$} & $0$ {\tiny $\pm 0$} & $6$ {\tiny $\pm 1$} & $4$ {\tiny $\pm 4$} & $0$ {\tiny $\pm 0$} & $5$ {\tiny $\pm 1$} & $\mathbf{83}$ {\tiny $\pm 10$} \\
\texttt{scene-play-singletask-task5-v0} & $0$ {\tiny $\pm 0$} & $0$ {\tiny $\pm 0$} & $0$ {\tiny $\pm 0$} & $0$ {\tiny $\pm 0$} & $0$ {\tiny $\pm 0$} & $0$ {\tiny $\pm 0$} & $0$ {\tiny $\pm 0$} & $0$ {\tiny $\pm 0$} & $0$ {\tiny $\pm 0$} & $0$ {\tiny $\pm 0$} & $\mathbf{2}$ {\tiny $\pm 2$} \\
\midrule
\texttt{puzzle-3x3-play-singletask-task1-v0} & $5$ {\tiny $\pm 2$} & $33$ {\tiny $\pm 6$} & $\mathbf{97}$ {\tiny $\pm 4$} & $52$ {\tiny $\pm 12$} & $89$ {\tiny $\pm 5$} & $\mathbf{97}$ {\tiny $\pm 2$} & $25$ {\tiny $\pm 9$} & $63$ {\tiny $\pm 19$} & $\mathbf{94}$ {\tiny $\pm 3$} & $90$ {\tiny $\pm 4$} & $87$ {\tiny $\pm 9$} \\
\texttt{puzzle-3x3-play-singletask-task2-v0} & $1$ {\tiny $\pm 1$} & $4$ {\tiny $\pm 3$} & $1$ {\tiny $\pm 1$} & $0$ {\tiny $\pm 1$} & $0$ {\tiny $\pm 1$} & $0$ {\tiny $\pm 0$} & $4$ {\tiny $\pm 2$} & $2$ {\tiny $\pm 2$} & $1$ {\tiny $\pm 2$} & $16$ {\tiny $\pm 5$} & $\mathbf{39}$ {\tiny $\pm 2$} \\
\texttt{puzzle-3x3-play-singletask-task3-v0} & $1$ {\tiny $\pm 1$} & $3$ {\tiny $\pm 2$} & $3$ {\tiny $\pm 1$} & $0$ {\tiny $\pm 0$} & $0$ {\tiny $\pm 0$} & $0$ {\tiny $\pm 0$} & $1$ {\tiny $\pm 0$} & $1$ {\tiny $\pm 1$} & $0$ {\tiny $\pm 0$} & $10$ {\tiny $\pm 3$} & $\mathbf{20}$ {\tiny $\pm 3$} \\
\texttt{puzzle-3x3-play-singletask-task4-v0 (*)} & $1$ {\tiny $\pm 1$} & $2$ {\tiny $\pm 1$} & $2$ {\tiny $\pm 1$} & $0$ {\tiny $\pm 0$} & $0$ {\tiny $\pm 0$} & $0$ {\tiny $\pm 0$} & $1$ {\tiny $\pm 1$} & $2$ {\tiny $\pm 2$} & $0$ {\tiny $\pm 0$} & $\mathbf{16}$ {\tiny $\pm 5$} & $10$ {\tiny $\pm 6$} \\
\texttt{puzzle-3x3-play-singletask-task5-v0} & $1$ {\tiny $\pm 0$} & $3$ {\tiny $\pm 2$} & $5$ {\tiny $\pm 3$} & $0$ {\tiny $\pm 0$} & $0$ {\tiny $\pm 0$} & $0$ {\tiny $\pm 0$} & $1$ {\tiny $\pm 1$} & $2$ {\tiny $\pm 2$} & $0$ {\tiny $\pm 0$} & $16$ {\tiny $\pm 3$} & $\mathbf{19}$ {\tiny $\pm 1$} \\
\midrule
\texttt{puzzle-4x4-play-singletask-task1-v0} & $1$ {\tiny $\pm 1$} & $12$ {\tiny $\pm 2$} & $26$ {\tiny $\pm 4$} & $48$ {\tiny $\pm 5$} & $24$ {\tiny $\pm 9$} & $44$ {\tiny $\pm 10$} & $1$ {\tiny $\pm 2$} & $32$ {\tiny $\pm 9$} & $49$ {\tiny $\pm 9$} & $34$ {\tiny $\pm 8$} & $\mathbf{72}$ {\tiny $\pm 8$} \\
\texttt{puzzle-4x4-play-singletask-task2-v0} & $0$ {\tiny $\pm 0$} & $7$ {\tiny $\pm 4$} & $12$ {\tiny $\pm 4$} & $14$ {\tiny $\pm 5$} & $0$ {\tiny $\pm 1$} & $0$ {\tiny $\pm 0$} & $0$ {\tiny $\pm 1$} & $5$ {\tiny $\pm 3$} & $4$ {\tiny $\pm 4$} & $\mathbf{16}$ {\tiny $\pm 5$} & $4$ {\tiny $\pm 2$} \\
\texttt{puzzle-4x4-play-singletask-task3-v0} & $0$ {\tiny $\pm 0$} & $9$ {\tiny $\pm 3$} & $15$ {\tiny $\pm 3$} & $34$ {\tiny $\pm 5$} & $21$ {\tiny $\pm 10$} & $29$ {\tiny $\pm 12$} & $1$ {\tiny $\pm 1$} & $20$ {\tiny $\pm 10$} & $\mathbf{50}$ {\tiny $\pm 14$} & $18$ {\tiny $\pm 5$} & $\mathbf{47}$ {\tiny $\pm 10$} \\
\texttt{puzzle-4x4-play-singletask-task4-v0 (*)} & $0$ {\tiny $\pm 0$} & $5$ {\tiny $\pm 2$} & $10$ {\tiny $\pm 3$} & $\mathbf{26}$ {\tiny $\pm 6$} & $7$ {\tiny $\pm 4$} & $1$ {\tiny $\pm 1$} & $0$ {\tiny $\pm 0$} & $5$ {\tiny $\pm 1$} & $21$ {\tiny $\pm 11$} & $11$ {\tiny $\pm 3$} & $10$ {\tiny $\pm 3$} \\
\texttt{puzzle-4x4-play-singletask-task5-v0} & $0$ {\tiny $\pm 0$} & $4$ {\tiny $\pm 1$} & $7$ {\tiny $\pm 3$} & $\mathbf{24}$ {\tiny $\pm 11$} & $1$ {\tiny $\pm 1$} & $0$ {\tiny $\pm 0$} & $0$ {\tiny $\pm 1$} & $4$ {\tiny $\pm 3$} & $2$ {\tiny $\pm 2$} & $7$ {\tiny $\pm 3$} & $2$ {\tiny $\pm 1$} \\
\midrule
\texttt{antmaze-umaze-v2} & $55$ & $77$ & $\mathbf{98}$ & $94$ & $\mathbf{97}$ & $66$ {\tiny $\pm 5$} & $90$ {\tiny $\pm 6$} & $94$ {\tiny $\pm 3$} & $92$ {\tiny $\pm 6$} & $\mathbf{96}$ {\tiny $\pm 2$} & $\mathbf{96}$ {\tiny $\pm 2$} \\
\texttt{antmaze-umaze-diverse-v2} & $47$ & $54$ & $84$ & $80$ & $82$ & $66$ {\tiny $\pm 11$} & $55$ {\tiny $\pm 7$} & $82$ {\tiny $\pm 9$} & $62$ {\tiny $\pm 12$} & $\mathbf{89}$ {\tiny $\pm 5$} & $\mathbf{86}$ {\tiny $\pm 9$} \\
\texttt{antmaze-medium-play-v2} & $0$ & $66$ & $\mathbf{90}$ & $84$ & $81$ & $49$ {\tiny $\pm 24$} & $52$ {\tiny $\pm 12$} & $77$ {\tiny $\pm 7$} & $56$ {\tiny $\pm 15$} & $78$ {\tiny $\pm 7$} & $81$ {\tiny $\pm 5$} \\
\texttt{antmaze-medium-diverse-v2} & $1$ & $74$ & $\mathbf{84}$ & $\mathbf{85}$ & $75$ & $0$ {\tiny $\pm 1$} & $44$ {\tiny $\pm 15$} & $77$ {\tiny $\pm 6$} & $60$ {\tiny $\pm 25$} & $71$ {\tiny $\pm 13$} & $75$ {\tiny $\pm 8$} \\
\texttt{antmaze-large-play-v2} & $0$ & $42$ & $52$ & $64$ & $54$ & $0$ {\tiny $\pm 0$} & $10$ {\tiny $\pm 6$} & $32$ {\tiny $\pm 21$} & $55$ {\tiny $\pm 9$} & $\mathbf{84}$ {\tiny $\pm 7$} & $\mathbf{83}$ {\tiny $\pm 6$} \\
\texttt{antmaze-large-diverse-v2} & $0$ & $30$ & $64$ & $68$ & $54$ & $0$ {\tiny $\pm 0$} & $16$ {\tiny $\pm 10$} & $20$ {\tiny $\pm 17$} & $64$ {\tiny $\pm 8$} & $\mathbf{83}$ {\tiny $\pm 4$} & $\mathbf{84}$ {\tiny $\pm 6$} \\
\midrule
\texttt{pen-human-v1} & $71$ & $78$ & $\mathbf{103}$ & $76$ {\tiny $\pm 10$} & $69$ {\tiny $\pm 7$} & $64$ {\tiny $\pm 8$} & $67$ {\tiny $\pm 5$} & $77$ {\tiny $\pm 7$} & $71$ {\tiny $\pm 12$} & $53$ {\tiny $\pm 6$} & $51$ {\tiny $\pm 10$} \\
\texttt{pen-cloned-v1} & $52$ & $83$ & $\mathbf{103}$ & $64$ {\tiny $\pm 7$} & $61$ {\tiny $\pm 7$} & $56$ {\tiny $\pm 10$} & $62$ {\tiny $\pm 10$} & $67$ {\tiny $\pm 9$} & $80$ {\tiny $\pm 11$} & $74$ {\tiny $\pm 11$} & $63$ {\tiny $\pm 9$} \\
\texttt{pen-expert-v1} & $110$ & $128$ & $\mathbf{152}$ & $140$ {\tiny $\pm 6$} & $134$ {\tiny $\pm 4$} & $103$ {\tiny $\pm 9$} & $118$ {\tiny $\pm 6$} & $119$ {\tiny $\pm 7$} & $139$ {\tiny $\pm 5$} & $142$ {\tiny $\pm 6$} & $\mathbf{145}$ {\tiny $\pm 6$} \\
\texttt{door-human-v1} & $2$ & $3$ & $-0$ & $6$ {\tiny $\pm 2$} & $3$ {\tiny $\pm 3$} & $5$ {\tiny $\pm 2$} & $2$ {\tiny $\pm 1$} & $4$ {\tiny $\pm 2$} & $\mathbf{7}$ {\tiny $\pm 2$} & $0$ {\tiny $\pm 0$} & $0$ {\tiny $\pm 0$}  \\
\texttt{door-cloned-v1} & $-0$ & $\mathbf{3}$ & $0$ & $0$ {\tiny $\pm 0$} & $0$ {\tiny $\pm 0$} & $1$ {\tiny $\pm 0$} & $0$ {\tiny $\pm 1$} & $0$ {\tiny $\pm 0$} & $2$ {\tiny $\pm 2$} & $2$ {\tiny $\pm 1$} & $0$ {\tiny $\pm 0$} \\
\texttt{door-expert-v1} & $\mathbf{105}$ & $\mathbf{107}$ & $\mathbf{106}$ & $\mathbf{105}$ {\tiny $\pm 1$} & $\mathbf{105}$ {\tiny $\pm 0$} & $98$ {\tiny $\pm 3$} & $\mathbf{103}$ {\tiny $\pm 1$} & $\mathbf{104}$ {\tiny $\pm 1$} & $\mathbf{104}$ {\tiny $\pm 2$} & $\mathbf{104}$ {\tiny $\pm 1$} & $\mathbf{105}$ {\tiny $\pm 1$} \\
\texttt{hammer-human-v1} & $\mathbf{3}$ & $2$ & $0$ & $2$ {\tiny $\pm 1$} & $1$ {\tiny $\pm 1$} & $2$ {\tiny $\pm 0$} & $2$ {\tiny $\pm 1$} & $2$ {\tiny $\pm 1$} & $\mathbf{3}$ {\tiny $\pm 1$} & $1$ {\tiny $\pm 1$} & $1$ {\tiny $\pm 1$} \\
\texttt{hammer-cloned-v1} & $1$ & $2$ & $5$ & $2$ {\tiny $\pm 1$} & $2$ {\tiny $\pm 1$} & $1$ {\tiny $\pm 1$} & $1$ {\tiny $\pm 0$} & $2$ {\tiny $\pm 1$} & $2$ {\tiny $\pm 1$} & $\mathbf{11}$ {\tiny $\pm 9$} & $1$ {\tiny $\pm 1$} \\
\texttt{hammer-expert-v1} & $127$ & $\mathbf{129}$ & $\mathbf{134}$ & $125$ {\tiny $\pm 4$} & $127$ {\tiny $\pm 0$} & $92$ {\tiny $\pm 11$} & $118$ {\tiny $\pm 3$} & $119$ {\tiny $\pm 9$} & $117$ {\tiny $\pm 9$} & $125$ {\tiny $\pm 3$} & $\mathbf{127}$ {\tiny $\pm 9$} \\
\texttt{relocate-human-v1} & $\mathbf{0}$ & $\mathbf{0}$ & $\mathbf{0}$ & $\mathbf{0}$ {\tiny $\pm 0$} & $\mathbf{0}$ {\tiny $\pm 0$} & $\mathbf{0}$ {\tiny $\pm 0$} & $\mathbf{0}$ {\tiny $\pm 0$} & $\mathbf{0}$ {\tiny $\pm 0$} & $\mathbf{0}$ {\tiny $\pm 0$} & $\mathbf{0}$ {\tiny $\pm 0$} & $\mathbf{0}$ {\tiny $\pm 0$} \\
\texttt{relocate-cloned-v1} & $-0$ & $0$ & $\mathbf{2}$ & $-0$ {\tiny $\pm 0$} & $-0$ {\tiny $\pm 0$} & $-0$ {\tiny $\pm 0$} & $-0$ {\tiny $\pm 0$} & $1$ {\tiny $\pm 1$} & $-0$ {\tiny $\pm 0$} & $-0$ {\tiny $\pm 0$} & $0$ {\tiny $\pm 0$} \\
\texttt{relocate-expert-v1} & $\mathbf{108}$ & $\mathbf{106}$ & $\mathbf{108}$ & $\mathbf{107}$ {\tiny $\pm 1$} & $\mathbf{106}$ {\tiny $\pm 2$} & $93$ {\tiny $\pm 6$} & $\mathbf{105}$ {\tiny $\pm 3$} & $\mathbf{105}$ {\tiny $\pm 2$} & $\mathbf{104}$ {\tiny $\pm 3$} & $\mathbf{107}$ {\tiny $\pm 1$} & $\mathbf{102}$ {\tiny $\pm 5$} \\
\midrule
\texttt{halfcheetah-medium-v2} & $42$ & $47$ & $\mathbf{64}$ & $51$ & $\mathbf{60}$ & $-$ & $-$ & $-$ & $55$ {\tiny $\pm 1$} & $59$ {\tiny $\pm 1$} & $59$ {\tiny $\pm 1$} \\
\texttt{halfcheetah-medium-replay-v2} & $36$ & $45$ & $\mathbf{51}$ & $46$ & $\mathbf{51}$ & $-$ & $-$ & $-$ & $44$ {\tiny $\pm 1$} & $50$ {\tiny $\pm 1$} & $\mathbf{53}$ {\tiny $\pm 1$} \\
\texttt{halfcheetah-medium-expert-v2} & $56$ & $96$ & $\mathbf{104}$ & $96$ & $92$ & $-$ & $-$ & $-$ & $79$ {\tiny $\pm 2$} & $89$ {\tiny $\pm 1$} & $96$ {\tiny $\pm 1$} \\
\texttt{hopper-medium-v2} & $54$ & $59$ & $\mathbf{102}$ & $65$ & $96$ & $-$ & $-$ & $-$ & $37$ {\tiny $\pm 6$} & $62$ {\tiny $\pm 6$} & $75$ {\tiny $\pm 6$} \\
\texttt{hopper-medium-replay-v2} & $30$ & $95$ & $95$ & $92$ & $\mathbf{102}$ & $-$ & $-$ & $-$ & $25$ {\tiny $\pm 1$} & $36$ {\tiny $\pm 6$} & $83$ {\tiny $\pm 6$} \\
\texttt{hopper-medium-expert-v2} & $52$ & $99$  & $\mathbf{110}$ & $\mathbf{109}$ & $100$ & $-$ & $-$ & $-$ & $69$ {\tiny $\pm 2$} & $71$ {\tiny $\pm 11$} & $\mathbf{110}$ {\tiny $\pm 3$} \\
\texttt{walker2d-medium-v2} & $63$ & $\mathbf{81}$ & $\mathbf{86}$ & $\mathbf{83}$ & $\mathbf{84}$ & $-$ & $-$ & $-$ & $35$ {\tiny $\pm 2$} & $69$ {\tiny $\pm 5$} & $\mathbf{81}$ {\tiny $\pm 1$} \\
\texttt{walker2d-medium-replay-v2} & $22$ & $73$ & $\mathbf{84}$ & $\mathbf{85}$ & $\mathbf{85}$ & $-$ & $-$ & $-$ & $13$ {\tiny $\pm 1$} & $39$ {\tiny $\pm 5$} & $58$ {\tiny $\pm 7$} \\
\texttt{walker2d-medium-expert-v2} & $99$ & $\mathbf{110}$ & $\mathbf{112}$ & $\mathbf{113}$ & $\mathbf{114}$ & $-$ & $-$ & $-$ & $94$ {\tiny $\pm 8$} & $96$ {\tiny $\pm 4$} & $\mathbf{110}$ {\tiny $\pm 1$} \\

\bottomrule
\end{tabular}
\end{threeparttable}
}
\end{table*}

\clearpage

\section{Algorithmic Details}
\label{app:appendix_algorithm}

\autoref{alg:drift} presents the complete pseudocode for the \ourMethod's drift loss. At each training step, a batch of transitions $(s, a^+, r, s')$ is sampled from the offline dataset $\mathcal{D}$. The behavioral regularization term is computed by generating $N$ action samples $\{\hat{a}_i\}$ from the actor network $f_\theta(s, \epsilon_i)$ with i.i.d. noise vectors $\epsilon_i \sim \mathcal{N}(0, I)$. These samples serve a dual role: they are the particles being transported toward the data distribution, and they simultaneously supply the empirical negatives used to compute the repulsive component of the drift field.

The drift vector $V_i$ for each particle decomposes into an attraction term $V^+_i$, which displaces $\hat{a}_i$ toward the single observed dataset action $a^+$, and a repulsion term $V^-_i$, which is a kernel-weighted average of displacements pointing \emph{toward} neighboring generated samples. Subtracting it from $V^+_i$ produces the net repulsive effect (\autoref{eq:drift}). The repulsion weights are normalized via a row-wise softmax over the negatives.

\begin{figure}
    \centering
    \includegraphics[width=1\linewidth]{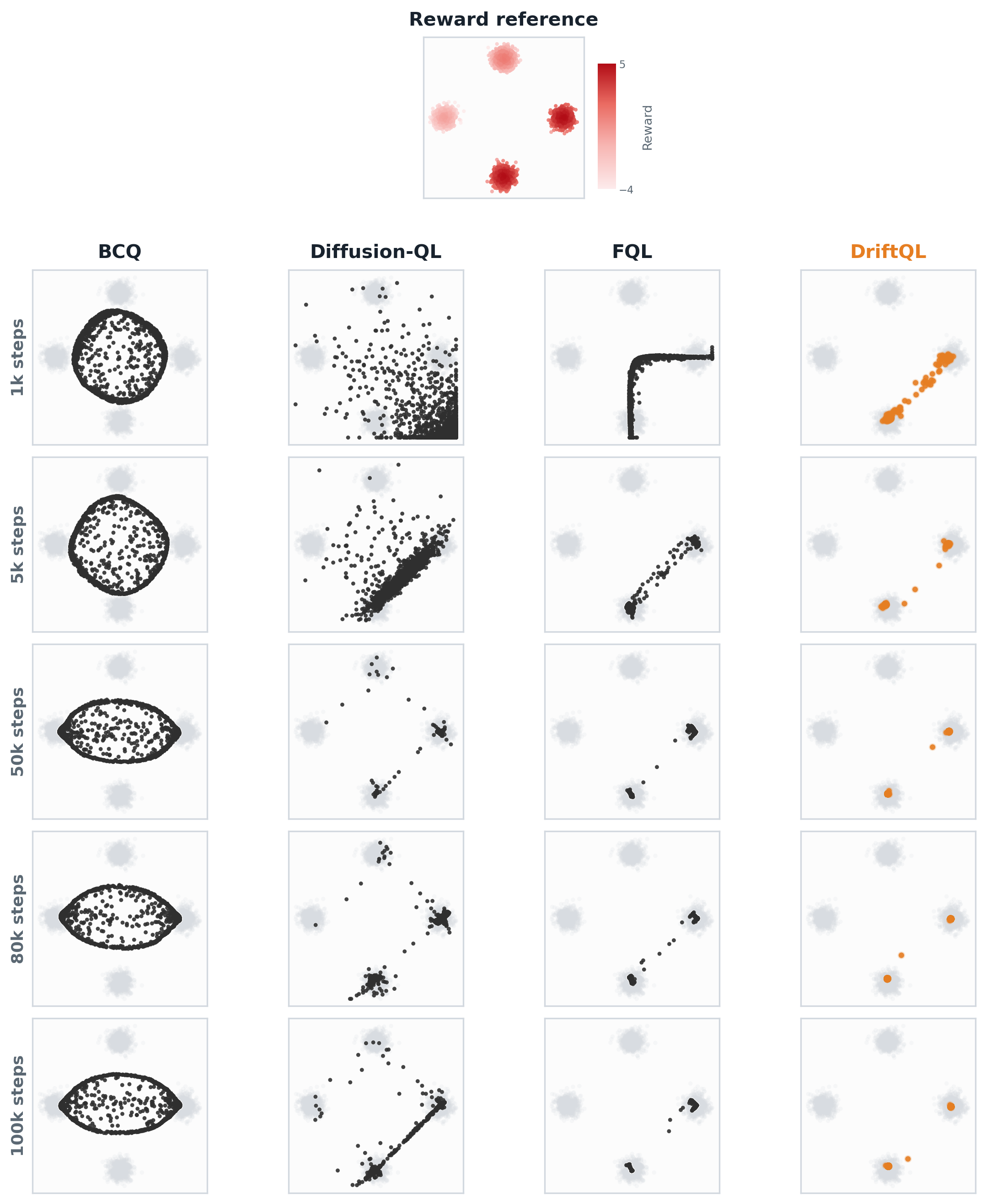}
    \caption{Detailed training checkpoints of the bandit example.}
    \label{fig:bandit_progression}
\end{figure}

The critic is updated via standard Bellman regression with a double ensemble and conservative minimum-Q targets to mitigate overestimation. The actor then jointly minimizes the drift loss $\mathcal{L}_{\text{drift}}$ and maximizes the mean/min ensemble Q-value, with the tradeoff governed by a single scalar $\alpha$. \autoref{fig:bandit_progression} shows detailed training checkpoints of the four-Gaussian bandit from \autoref{fig:intuition_bandit}, tracking how each method's generated actions evolve over training.

\input{assets/algo}

\clearpage

\clearpage

\section{\ourMethod vs. Original Drifting}\label{app:contragen}

This ablation isolates the effect of the drift computation. Both variants share the same state-conditioned actor, critic, Q-maximization term, number of generated samples, and single dataset action $a^+$ per state. Only the drift regularizer differs.

The original drifting implementation~\citep{driftmodels} pools generated samples, positives, and optional negatives into a joint set, computes symmetrized affinities, cross-weights positive and negative terms, aggregates over several kernel radii, and normalizes the resulting force. \ourMethod instead uses the decomposition from \autoref{sec:driftql}: the dataset action serves as a direct anchor $V^+(\hat a_i)=a^+-\hat a_i$, while the remaining generated actions estimate the current policy through $V^-(\hat a_i)=\sum_{j\neq i}w^-_{ij}(\hat a_j-\hat a_i)$.

This distinction matters because the drift term is not a standalone training signal: it is paired with a critic that pushes the actor toward high predicted value. As \citet{lai2026unified} noted, finite-sample choices such as softmax normalization, positive/negative balance, batch-dependent scaling, and temperature aggregation reshape the effective transport field. The original computation introduces batch- and kernel-dependent rescalings that interact with the Q gradient in hard-to-predict ways. Our formulation removes these, so the behavioral correction is controlled directly by $\alpha$ in \autoref{eq:actor_loss}.

\begin{figure}[t]
    \centering
    \includegraphics[width=0.5\linewidth]{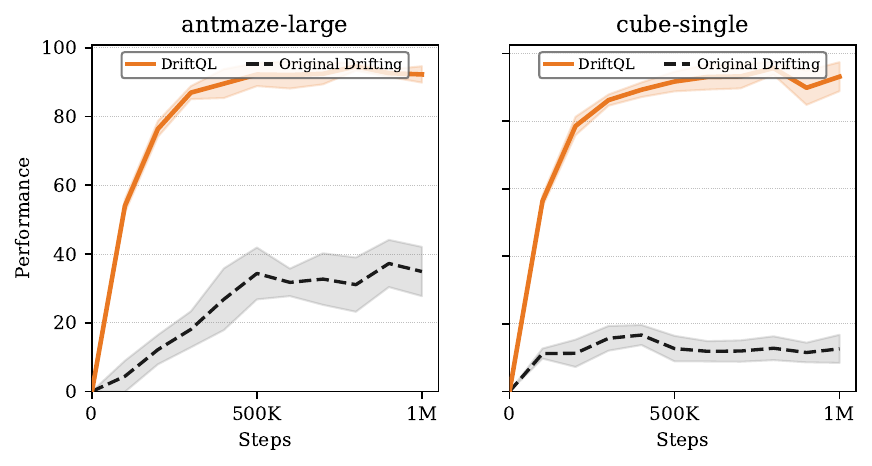}
    \caption{\textbf{DriftQL vs. original drifting in the offline RL setting.} Performance on OGBench AntMaze Large and Cube Single (default) over 1M training steps, averaged across three seeds. Shaded regions denote standard deviation.}
    \label{fig:driftqlvsoriginaldrifting}
\end{figure}

\autoref{fig:driftqlvsoriginaldrifting} shows that our formulation is more suited when combined with Q maximization. We swept over multiple values of $\alpha$ and the kernel temperature for the original drift variant and found significant instability across runs, with performance highly sensitive to both hyperparameters. \ourMethod keeps the same attraction-repulsion structure but uses a drift field better suited to low-dimensional, state-conditioned action spaces.

\subsection{Anti-symmetry: the critic absorbs the symmetrization machinery}
\label{app:antisymmetry}

\autoref{sec:driftql} argues that in offline RL the value gradient, not the symmetrization machinery of \citet{driftmodels}, is what anchors the actor, so the machinery can be dropped. We test this prediction directly with a destructive anti-symmetry probe. The attraction $V^+$ and repulsion $V^-$ are anti-symmetric by construction: at the population fixed point $p=q$ the two cancel. We deliberately break this balance by rescaling the attraction weight by a multiplier $\rho \in [0.01, 50]$ relative to the repulsion (default $\rho=1$). For $\rho \ne 1$ the field no longer cancels at $p=q$, the finite-sample analogue of the residual ambiguity that \citet{lai2026unified} (Thm. 2) flag at the population level. We compare \textsc{pure-drift} (drift only, no $Q$) against \textsc{full-\ourMethod} (drift $+Q$), logging divergence when actions exit a fixed bounding box.

\begin{table}[t]
\centering
\small
\caption{Anti-symmetry stress test. $\rho$ rescales the attraction weight relative to repulsion. \textsc{pure-drift} diverges or fails to concentrate for $\rho \le 1$, recovering only when attraction dominates ($\rho \gtrsim 5$). \textsc{full-\ourMethod} is essentially flat across more than three orders of magnitude.}
\begin{tabular}{lcc}
\toprule
$V^+$ multiplier $\rho$ & \textsc{pure-drift} (Y-var) & \textsc{full-\ourMethod} (Y-var) \\
\midrule
$0.01$  & \textit{DIVERGED} & $0.0011$ \\
$0.05$  & \textit{DIVERGED} & $0.0009$ \\
$0.10$  & \textit{DIVERGED} & $0.0006$ \\
$0.20$  & \textit{DIVERGED} & $0.0006$ \\
$0.50$  & $0.1523$        & $0.0006$ \\
$1.00$  & $0.0673$        & $0.0006$ \\
$2.00$  & $0.0152$        & $0.0010$ \\
$5.00$  & $0.0009$        & $0.0007$ \\
$10.00$ & $0.0006$        & $0.0006$ \\
$20.00$ & $0.0006$        & $0.0006$ \\
$50.00$ & $0.0005$        & $0.0006$ \\
\bottomrule
\end{tabular}
\label{tab:antisym}
\end{table}

\begin{figure}[t]
\centering
\includegraphics[width=0.5\linewidth]{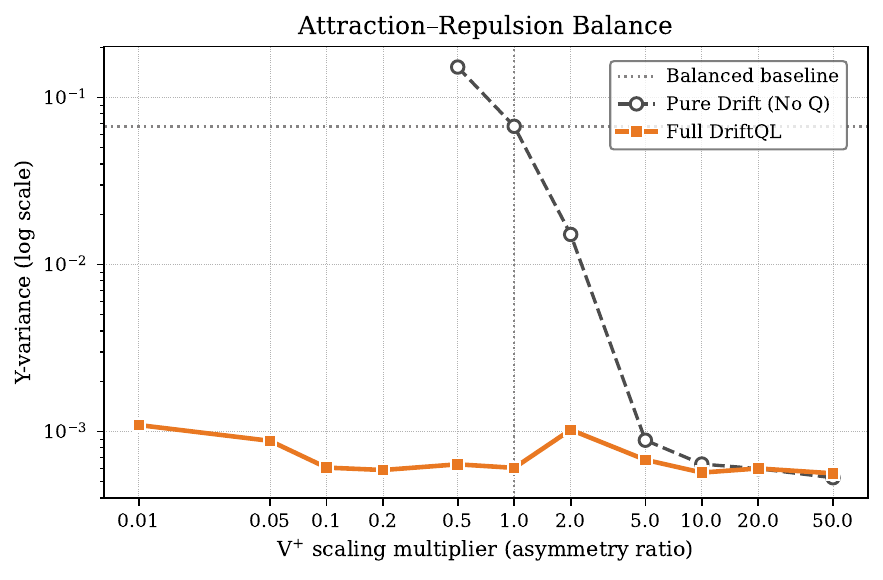}
\caption{\textbf{Anti-symmetry stress test.} Final Y-variance vs. attraction-rescaling multiplier $\rho$, log--log. The pure-drift baseline is unstable across most of the range. It diverges for $\rho \le 0.2$ (off-plot) and only stabilizes once attraction strongly dominates repulsion. \textsc{full-\ourMethod} is essentially constant across the entire sweep. The critic gradient supplies a separate value-driven pull that keeps the actor stable even when the drift field is severely mis-balanced.}
\label{fig:antisym}
\end{figure}

\autoref{tab:antisym} and \autoref{fig:antisym} confirm the prediction. \textsc{pure-drift} is fragile: a small under-weighting of attraction ($\rho \le 0.2$) makes the actor diverge, because $V^-$ pushes samples apart with nothing to anchor them, exactly the failure the original machinery is built to prevent. \textsc{full-\ourMethod}, by contrast, is essentially flat across more than three orders of magnitude. The $Q$ gradient supplies an independent, value-driven pull toward high-value support, so even when the drift's attraction is artificially crippled the critic keeps the actor from drifting away. This is the empirical content of the claim in \autoref{sec:driftql}: in offline RL the critic already supplies the constraint that cross-weighting, multi-temperature pooling, and RMS rescaling supply in the unconditional setting, so \ourMethod can drop them.

\clearpage

\section{Network Architectures}
\label{app:architecture}
To effectively parameterize the drift policy and value functions, we utilize MLPs optimized for continuous control tasks. The architecture consists of a stochastic generator (Actor) and a double-ensemble Q-network (Critic).

As detailed below, the Critic network applies Layer Normalization after each linear transformation to stabilize value estimation and prevent early divergence during offline training. Conversely, the Actor network maps the concatenated state and noise inputs through a deep network without Layer Normalization, culminating in a bounded action output. Both networks rely on a wide and deep structure (4 hidden layers of 512 units) to ensure sufficient expressivity when modeling multimodal behavioral distributions and complex value landscapes.
\vspace{2em}

\begin{tcolorbox}[title={\textbf{\ourMethod Network Architectures}},
  colback=myorange!5,
  colframe=myorange!60,
  coltitle=black,
  fonttitle=\bfseries\small,
  rounded corners,
  arc=3pt,
  boxrule=0.5pt
]
\label{box:architecture}
\footnotesize
\textbf{Inputs:}
\begin{itemize}
    \item Observation $s \in \mathbb{R}^S$
    \item Action $a \in \mathbb{R}^A$ (Dataset action or generated sample)
    \item Noise $\epsilon \sim \mathcal{N}(0, I) \in \mathbb{R}^A$
\end{itemize}
\vspace{1em}

\textbf{1. Critic (Value) Network $Q_\phi(s, a)$:}
\begin{verbatim}
# Double ensemble architecture (num_ensembles=2)
# Applies LayerNorm to stabilize value bootstrapping
concat = Concatenate([s, a]) # dim: S + A
x_q = Linear(S + A, 512) -> LayerNorm -> GELU
x_q = Linear(512, 512)   -> LayerNorm -> GELU
x_q = Linear(512, 512)   -> LayerNorm -> GELU
x_q = Linear(512, 512)   -> LayerNorm -> GELU
Q   = Linear(512, 1)
\end{verbatim}
\vspace{1em}

\textbf{2. Actor (Generator) Network $f_\theta(s, z)$:}
\begin{verbatim}
# Generates pushforward distribution via noise z
# No LayerNorm applied (actor_layer_norm=False)
concat = Concatenate([s, z]) # dim: S + A
x_a = Linear(S + A, 512) -> GELU
x_a = Linear(512, 512)   -> GELU
x_a = Linear(512, 512)   -> GELU
x_a = Linear(512, 512)   -> GELU
a_raw = Linear(512, A)
action = Clip(a_raw, -1.0, 1.0)
\end{verbatim}
\end{tcolorbox}

\clearpage
\section{Limitations} \label{app:limitations}

We discuss the limitations of \ourMethod along three axes: the scope of theoretical support, computational considerations, and the scope of our empirical evaluation.


\textbf{Theoretical guarantees only partially transfer.} The reverse-Fisher interpretation of \citet{lai2026unified} is established for the original unconditional drifting objective with coupled cross-normalization between attraction and repulsion. Our construction departs from this in three ways: we condition the drift field on state, we normalize attraction and repulsion independently, and we operate in the single-positive regime. The motivation for these changes is empirical and structural (\autoref{sec:driftql}), not theoretical, and we do not claim that the $O(\tau^{4})$ smoothed reverse-Fisher rate or the $O(D^{-(1+2a)})$ minimizer discrepancy bound carry over unchanged. Extending the analysis of \citet{lai2026unified} to conditional, single-positive drift fields optimized jointly with a learned critic is an open problem.

\textbf{Training-time cost offsets the inference-time gain.} Although \ourMethod requires only a single forward pass at evaluation, each training step draws $N$ generator samples per state and computes an $N \times N$ pairwise kernel (\autoref{alg:drift}). However, wall-clock training time is close to baselines. We found $N = 32$ to be a reasonable operating point, but this trades sample diversity for throughput and has not been tuned for wall-clock parity against baselines.

\textbf{Scope of action spaces.} The drift field as formulated assumes a continuous, Euclidean, box-bounded action space. We rely on the $\sqrt{d_a}$-scaled Euclidean kernel and a terminal $\mathrm{clip}(\cdot, -1, 1)$. Other types of action spaces are not supported without a redesign of both the kernel and the clipping step.

\clearpage
\section{Hyperparameters}\label{app:hyperparams}

In this section, we detail the environment-specific hyperparameters used for evaluating \ourMethod. To prioritize simplicity and minimize tuning overhead, the majority of our algorithmic configurations, such as network architectures, learning rates, optimization parameters, and the number of generated drift samples ($N$), remain entirely fixed across all benchmark tasks, as defined in our core implementation. \autoref{table:default_hyperparams} lists these fixed default hyperparameters shared across all environments.

\begin{table}[ht]
\centering
\footnotesize
\setlength{\tabcolsep}{4pt}
\caption{Default hyperparameters for \ourMethod fixed across all tasks.}
\label{table:default_hyperparams}
\begin{tabular}{@{}ll@{\hspace{16pt}}ll@{}}
\toprule
\textbf{Hyperparameter} & \textbf{Value} & \multicolumn{2}{@{}l}{\textit{\textbf{Drift Field Configurations}}}  \textbf{Value} \\ \midrule
Optimizer & Adam & Number of generated samples ($N$) & 32 \\
Learning rate & $3 \times 10^{-4}$ & Drift batch size & 256 \\
Minibatch size & 256 & Drift step size ($\eta$) & 1.0 \\
Actor MLP dimensions & $[512, 512, 512, 512]$ & Distance dimensionality scaling & True \\
Critic MLP dimensions & $[512, 512, 512, 512]$ & Drift normalize & False \\
Nonlinearity & GELU & Drift $\epsilon$ & $10^{-12}$ \\
Actor layer normalization & False & & \\
Critic layer normalization & True & & \\
Target network update rate ($\tau$) & 0.005 & & \\
Normalize Q-loss & False & & \\ \bottomrule
\end{tabular}
\end{table}

However, to account for the highly varied reward scales, dataset suboptimalities, and dimensionalities across the OGBench and D4RL suites, we tune three hyperparameters for each environment category. These parameters are:

\begin{itemize}
    \item \textbf{\texttt{drift\_temp}} ($\tau$): The temperature scalar controlling the sharpness of the similarity kernel in the drift field computation.
    \item \textbf{$\alpha$}: The trade-off coefficient balancing the drifting behavioral cloning loss against the Q-maximization objective.
    \item \textbf{\texttt{kernel}}: The distance kernel used to compute the drift field logits (either \texttt{laplace} or \texttt{gaussian}).
\end{itemize}

\autoref{table:hyperparams} summarizes the exact hyperparameter configurations used for each task domain. For OGBench environments, we follow the standard evaluation protocol by tuning these parameters on the default task (indicated by (*)) and applying the best configuration to the remaining four tasks within that environment suite.

\begin{table}[t]
\centering
\footnotesize
\setlength{\tabcolsep}{4pt}
\caption{Task-specific hyperparameters for \ourMethod.}
\label{table:hyperparams}
\begin{tabular}{@{}lccc@{}}
\toprule
\textbf{Task Domain} & \textbf{\texttt{drift\_temp}} ($\tau$) & \textbf{$\alpha$} & \textbf{\texttt{kernel}} \\ \midrule
\multicolumn{4}{@{}l}{\textit{\textbf{OGBench Locomotion}}} \\ \midrule
\texttt{antmaze-large-navigate-*}       & 0.5 & 10 &  \\
\texttt{antmaze-giant-navigate-*}       & 0.2 & 10 &   \\
\texttt{humanoidmaze-medium-navigate-*} & 0.5 & 65 & laplace \\
\texttt{humanoidmaze-large-navigate-*}  & 0.2 & 32 &   \\
\texttt{antsoccer-arena-navigate-*}     & 0.5 & 10 &   \\ \midrule
\multicolumn{4}{@{}l}{\textit{\textbf{OGBench Manipulation}}} \\ \midrule
\texttt{cube-single-play-*}             & 0.02 & 60 &   \\
\texttt{cube-double-play-*}             & 0.2 & 100 &  \\
\texttt{scene-play-*}                   & 0.2 & 250 &  gaussian \\
\texttt{puzzle-3x3-play-*}              & 0.5 & 50 &   \\
\texttt{puzzle-4x4-play-*}              & 0.8 & 300 &   \\ \midrule
\multicolumn{4}{@{}l}{\textit{\textbf{D4RL Antmaze}}} \\ \midrule
\texttt{antmaze-umaze-v2}               &  & 15 &  laplace \\
\texttt{antmaze-umaze-diverse-v2}       &  & 12 &  laplace \\
\texttt{antmaze-medium-play-v2}         &  & 5 &  gaussian \\
\texttt{antmaze-medium-diverse-v2}      & 0.5 & 8 &  gaussian \\
\texttt{antmaze-large-play-v2}          &  & 3 & gaussian \\
\texttt{antmaze-large-diverse-v2}       &  & 5 &  gaussian \\ \midrule
\multicolumn{4}{@{}l}{\textit{\textbf{D4RL Adroit}}} \\ \midrule
\texttt{pen-cloned-v1}                       & 0.05 & 1500 & gaussian \\
\texttt{pen-expert-v1}                       & 0.9 & 2000 & laplace \\
\texttt{pen-human-v1}                       & 0.05 & 2000 & laplace \\
\texttt{door-*-v1}                      & 0.2 & 4500 & laplace \\
\texttt{hammer-*-v1}                    & 0.05 & 2500 & laplace \\
\texttt{relocate-*-v1}                  & 0.2 & 5000 & laplace \\ \midrule
\multicolumn{4}{@{}l}{\textit{\textbf{D4RL Locomotion}}} \\ \midrule
\texttt{halfcheetah-medium-expert-v2}               & 0.5 & 300 & laplace \\
\texttt{halfcheetah-medium-replay-v2}               & 0.5 & 10 & laplace \\
\texttt{halfcheetah-medium-v2}               & 0.5 & 3 & laplace \\
\texttt{hopper-medium-expert-v2}                    & 0.1 & 600 &  gaussian \\
\texttt{hopper-medium-replay-v2}                    & 0.1 & 100 & gaussian \\
\texttt{hopper-medium-v2}                    & 0.1 & 100 & gaussian \\
\texttt{walker2d-medium-expert-v2}                  & 0.1 & 1000 & laplace \\
\texttt{walker2d-medium-replay-v2}                  & 0.1 & 300 & laplace \\
\texttt{walker2d-medium-v2}                         & 0.1 & 1000 & laplace \\
\bottomrule
\end{tabular}
\end{table}


%% file: assets/algo.tex
\tikzexternaldisable
\begin{algorithm*}[ht]
\small
\caption{Drift Field Computation (actor loss, per training step)}
\label{alg:drift}
\begin{algorithmic}[1]
\BeginBox[fill=gray!20]
\State Require:
\Comment{Setup}
\Statex \quad State $s$, dataset action $a^+$, policy $f_\theta$, temperature $\tau$, kernel type $\in \{\text{Laplace}, \text{Gaussian}\}$
\EndBox

\BeginBox[fill=gray!10]
\State draw $\epsilon_1,\ldots,\epsilon_{N_\text{gen}} \sim \mathcal{N}(0,I)$
\Comment{Generate policy samples}
\State $\hat a_i \gets \operatorname{clip}\!\big(f_\theta(s,\epsilon_i), -1, 1\big)$ for each $i$
\EndBox

\BeginBox[fill=gray!20]
\State $d^-_{ik} \gets \|\hat a_i-\hat a_k\|_2/\sqrt{d_a}$ for each $i\neq k$
\Comment{Generated-action distances}
\State $d^-_{ii} \gets +\infty$
\Comment{Mask self-repulsion}
\EndBox

\BeginBox[fill=gray!10]
\State $\ell^-_{ik} \gets
\begin{cases}
-d^-_{ik}/\tau, & \text{Laplace},\\
-(d^-_{ik})^2/(2\tau^2), & \text{Gaussian}
\end{cases}$
\Comment{Kernel logits}
\State $w^-_{ik} \gets
\frac{\exp(\ell^-_{ik})}
{\sum_{k'\neq i}\exp(\ell^-_{ik'})}$
for $k\neq i$, and $w^-_{ii}\gets 0$
\Comment{Softmax over other generated actions}
\EndBox

\BeginBox[fill=gray!20]
\State $V^+_i \gets a^+ - \hat a_i$
\Comment{Attraction toward the dataset action}
\State $V^-_i \gets \sum_{k\neq i} w^-_{ik}(\hat a_k-\hat a_i)$
\Comment{Model-side mean-shift term}
\State $V_i \gets V^+_i - V^-_i$
\Comment{Full drift}
\EndBox

\BeginBox[fill=gray!10]
\State $\text{target}_i \gets
\operatorname{sg}\!\left(
\operatorname{clip}(\hat a_i+V_i,-1,1)
\right)$
\Comment{Stop-gradient drifted target}
\State $\mathcal{L}_\text{drift}
\gets
\frac{1}{N_\text{gen}}
\sum_i
\|\hat a_i-\text{target}_i\|_2^2$
\Comment{Drift loss}
\EndBox
\end{algorithmic}
\end{algorithm*}
\tikzexternalenable

%% file: references.bib
@article{lai2026unified,
  title={A Unified View of Drifting and Score-Based Models},
  author={Lai, Chieh-Hsin and Nguyen, Bac and Murata, Naoki and Takida, Yuhta and Uesaka, Toshimitsu and Mitsufuji, Yuki and Ermon, Stefano and Tao, Molei},
  journal={arXiv preprint arXiv:2603.07514},
  year={2026}
}

@article{yang2022rorl,
  title={{RORL}: Robust offline reinforcement learning via conservative smoothing},
  author={Yang, Rui and Bai, Chenjia and Ma, Xiaoteng and Wang, Zhaoran and Zhang, Chongjie and Han, Lei},
  journal={Advances in neural information processing systems},
  volume={35},
  pages={23851--23866},
  year={2022}
}

@article{ghasemipour2022so,
  title={Why so pessimistic? {E}stimating uncertainties for offline {RL} through ensembles, and why their independence matters},
  author={Ghasemipour, Kamyar and Gu, Shixiang Shane and Nachum, Ofir},
  journal={Advances in Neural Information Processing Systems},
  volume={35},
  pages={18267--18281},
  year={2022}
}

@inproceedings{
an2021uncertaintybased,
title={Uncertainty-Based Offline Reinforcement Learning with Diversified Q-Ensemble},
author={Gaon An and Seungyong Moon and Jang-Hyun Kim and Hyun Oh Song},
booktitle={Advances in Neural Information Processing Systems},
editor={A. Beygelzimer and Y. Dauphin and P. Liang and J. Wortman Vaughan},
year={2021},
url={https://openreview.net/forum?id=ZUvaSolQZh3}
}

@article{tarasov2023revisiting,
  title={Revisiting the minimalist approach to offline reinforcement learning},
  author={Tarasov, Denis and Kurenkov, Vladislav and Nikulin, Alexander and Kolesnikov, Sergey},
  journal={Advances in Neural Information Processing Systems},
  volume={36},
  pages={11592--11620},
  year={2023}
}

@article{tarasov2023corl,
  title={{CORL}: Research-oriented deep offline reinforcement learning library},
  author={Tarasov, Denis and Nikulin, Alexander and Akimov, Dmitry and Kurenkov, Vladislav and Kolesnikov, Sergey},
  journal={Advances in Neural Information Processing Systems},
  volume={36},
  pages={30997--31020},
  year={2023}
}

@inproceedings{
chae2026flow,
title={Flow Actor-Critic for Offline Reinforcement Learning},
author={Jongseong Chae and Jongeui Park and Yongjae Shin and Gyeongmin Kim and Seungyul Han and Youngchul Sung},
booktitle={The Fourteenth International Conference on Learning Representations},
year={2026},
url={https://openreview.net/forum?id=wuncwN7iZN}
}

@inproceedings{
ding2024consistency,
title={Consistency Models as a Rich and Efficient Policy Class for Reinforcement Learning},
author={Zihan Ding and Chi Jin},
booktitle={The Twelfth International Conference on Learning Representations},
year={2024},
url={https://openreview.net/forum?id=v8jdwkUNXb}
}

@inproceedings{
chen2024score,
title={Score Regularized Policy Optimization through Diffusion Behavior},
author={Huayu Chen and Cheng Lu and Zhengyi Wang and Hang Su and Jun Zhu},
booktitle={The Twelfth International Conference on Learning Representations},
year={2024},
url={https://openreview.net/forum?id=xCRr9DrolJ}
}

@book{sutton1998reinforcement,
  title={Reinforcement learning: An introduction},
  author={Sutton, Richard S and Barto, Andrew G and others},
  volume={1},
  year={1998},
  publisher={MIT press Cambridge}
}

@article{fujimoto2021td3bc,
  title={A minimalist approach to offline reinforcement learning},
  author={Fujimoto, Scott and Gu, Shixiang Shane},
  journal={Advances in neural information processing systems},
  year={2021}
}

@article{kumar2020cql,
  title={Conservative {Q}-learning for offline reinforcement learning},
  author={Kumar, Aviral and Zhou, Aurick and Tucker, George and Levine, Sergey},
  journal={Advances in neural information processing systems},
  year={2020}
}

@article{fu2021d4rl,
  title={D4{R}{L}: Datasets for deep data-driven reinforcement learning},
  author={Fu, Justin and Kumar, Aviral and Nachum, Ofir and Tucker, George and Levine, Sergey},
  journal={arXiv preprint arXiv:2004.07219},
  year={2020}
}

@inproceedings{
gao2025behaviorregularized,
title={Behavior-Regularized Diffusion Policy Optimization for Offline Reinforcement Learning},
author={Chen-Xiao Gao and Chenyang Wu and Mingjun Cao and Chenjun Xiao and Yang Yu and Zongzhang Zhang},
booktitle={Forty-second International Conference on Machine Learning},
year={2025},
}

@article{danesh2025safe,
  title={Safe Domain Randomization via Uncertainty-Aware Out-of-Distribution Detection and Policy Adaptation},
  author={Danesh, Mohamad H and Wabartha, Maxime and Wu, Stanley and Pineau, Joelle and Lin, Hsiu-Chin},
  journal={arXiv preprint arXiv:2507.06111},
  year={2025}
}

@article{koirala2025flow,
  title={Flow-Based Single-Step Completion for Efficient and Expressive Policy Learning},
  author={Koirala, Prajwal and Fleming, Cody},
  journal={arXiv preprint arXiv:2506.21427},
  year={2025}
}

@article{alles2025flowq, title={Flow{Q}: Energy-guided flow policies for offline reinforcement learning}, author={Alles, Marvin and Chen, Nutan and van der Smagt, Patrick and Cseke, Botond}, journal={arXiv preprint arXiv:2505.14139}, year={2025} }

@inproceedings{
abyaneh2026contractive,
title={Contractive Diffusion Policies: Robust Action Diffusion via Contractive Score-Based Sampling with Differential Equations},
author={Amin Abyaneh and Charlotte Morissette and Mohamad H. Danesh and Anas Houssaini and David Meger and Gregory Dudek and Hsiu-Chin Lin},
booktitle={The Fourteenth International Conference on Learning Representations},
year={2026},
url={https://openreview.net/forum?id=iKJbmx1iuQ}
}

@inproceedings{janner2022planning_diffusion,
  title={Planning with Diffusion for Flexible Behavior Synthesis},
  author={Janner, Michael and Du, Yilun and Tenenbaum, Joshua and Levine, Sergey},
  booktitle={International Conference on Machine Learning},
  year={2022},
}

@article{hansen2023idql,
  title={{IDQL}: Implicit {Q}-learning as an actor-critic method with diffusion policies},
  author={Hansen-Estruch, Philippe and Kostrikov, Ilya and Janner, Michael and Kuba, Jakub Grudzien and Levine, Sergey},
  journal={arXiv preprint arXiv:2304.10573},
  year={2023}
}

@inproceedings{
diffusionQL,
title={Diffusion Policies as an Expressive Policy Class for Offline Reinforcement Learning},
author={Zhendong Wang and Jonathan J Hunt and Mingyuan Zhou},
booktitle={The Eleventh International Conference on Learning Representations },
year={2023},
}

@inproceedings{
kang2023efficientDP,
title={Efficient Diffusion Policies For Offline Reinforcement Learning},
author={Bingyi Kang and Xiao Ma and Chao Du and Tianyu Pang and Shuicheng YAN},
booktitle={Thirty-seventh Conference on Neural Information Processing Systems},
year={2023},
}

@inproceedings{
zhang2026reform,
title={Re{FORM}: Reflected Flows for On-support Offline {RL} via Noise Manipulation},
author={Songyuan Zhang and Oswin So and H M Sabbir Ahmad and Eric Yang Yu and Matthew Cleaveland and Mitchell Black and Chuchu Fan},
booktitle={The Fourteenth International Conference on Learning Representations},
year={2026},
url={https://openreview.net/forum?id=YvFsyRReeN}
}

@inproceedings{
nguyen2026onestep,
title={One-Step Flow {Q}-Learning: Addressing the Diffusion Policy Bottleneck in Offline Reinforcement Learning},
author={Thanh Xuan Nguyen and Chang D. Yoo},
booktitle={The Fourteenth International Conference on Learning Representations},
year={2026},
url={https://openreview.net/forum?id=60VgwdzxDM}
}

@article{ho2020ddpm_diffusion,
  title={Denoising diffusion probabilistic models},
  author={Ho, Jonathan and Jain, Ajay and Abbeel, Pieter},
  journal={Advances in neural information processing systems},
  year={2020}
}

@inproceedings{sohl2015deep_diffusion_original,
  title={Deep unsupervised learning using nonequilibrium thermodynamics},
  author={Sohl-Dickstein, Jascha and Weiss, Eric and Maheswaranathan, Niru and Ganguli, Surya},
  booktitle={International conference on machine learning},
  year={2015},
}

@inproceedings{ogbench_park2025,
  title={{OGBench}: Benchmarking Offline Goal-Conditioned {RL}},
  author={Park, Seohong and Frans, Kevin and Eysenbach, Benjamin and Levine, Sergey},
  booktitle={International Conference on Learning Representations (ICLR)},
  year={2025},
}

@inproceedings{fql_park2025,
  title={Flow {Q}-Learning},
  author={Seohong Park and Qiyang Li and Sergey Levine},
  booktitle={International Conference on Machine Learning (ICML)},
  year={2025},
}

@misc{driftmodels,
      title={Generative Modeling via Drifting}, 
      author={Mingyang Deng and He Li and Tianhong Li and Yilun Du and Kaiming He},
      year={2026},
      eprint={2602.04770},
      archivePrefix={arXiv},
      primaryClass={cs.LG},
      url={https://arxiv.org/abs/2602.04770}, 
}

@article{Lipman2022FlowMF,
  title={Flow Matching for Generative Modeling},
  author={Yaron Lipman and Ricky T. Q. Chen and Heli Ben-Hamu and Maximilian Nickel and Matt Le},
  journal={ArXiv},
  year={2022},
  volume={abs/2210.02747},
  url={https://api.semanticscholar.org/CorpusID:252734897}
}

@article{kumar2019bear,
  author       = {Aviral Kumar and
                  Justin Fu and
                  George Tucker and
                  Sergey Levine},
  title        = {Stabilizing Off-Policy {Q}-Learning via Bootstrapping Error Reduction},
  journal      = {CoRR},
  volume       = {abs/1906.00949},
  year         = {2019},
  url          = {http://arxiv.org/abs/1906.00949},
  eprinttype    = {arXiv},
  eprint       = {1906.00949},
  bibsource    = {dblp computer science bibliography, https://dblp.org}
}

@article{kostrikov2021iql,
  author       = {Ilya Kostrikov and
                  Ashvin Nair and
                  Sergey Levine},
  title        = {Offline Reinforcement Learning with Implicit {Q}-Learning},
  journal      = {CoRR},
  volume       = {abs/2110.06169},
  year         = {2021},
  url          = {https://arxiv.org/abs/2110.06169},
  eprinttype    = {arXiv},
  eprint       = {2110.06169},
  bibsource    = {dblp computer science bibliography, https://dblp.org}
}

@inproceedings{fujimoto2019bcq,
  title={Off-Policy Deep Reinforcement Learning without Exploration},
  author={Fujimoto, Scott and Meger, David and Precup, Doina},
  booktitle={International Conference on Machine Learning},
  pages={2052--2062},
  year={2019}
}

@inproceedings{fujimoto2018addressing,
  title={Addressing function approximation error in actor-critic methods},
  author={Fujimoto, Scott and Hoof, Herke and Meger, David},
  booktitle={International conference on machine learning},
  pages={1587--1596},
  year={2018},
  organization={PMLR}
}
